\documentclass[10pt,journal,compsoc]{IEEEtran}

\usepackage{mathrsfs}
\usepackage{color}
\usepackage{bm}
\usepackage{amsfonts}
\usepackage{CJK}
\usepackage{amsmath}
\usepackage{algorithm}
\usepackage{indentfirst}
\usepackage{longtable}
\usepackage{supertabular}
\usepackage{multirow}
\usepackage{graphicx}
\usepackage{subfig}
\usepackage{stfloats}
\usepackage{verbatim}
\usepackage{algpseudocode}
\usepackage{url}
\usepackage{booktabs}
\usepackage{lipsum}
\usepackage{siunitx}
\usepackage{rotating}
\usepackage{caption}
\usepackage{fmtcount}

\def\BF{\textbf}

\def\etal{et al.}
\def\DXYB{\textcolor{black}}

\usepackage{array}
\newcolumntype{I}{!{\vrule width 3pt}}
\newlength\savewidth

\ifCLASSOPTIONcompsoc
  \usepackage[nocompress]{cite}
\else
  \usepackage{cite}

\fi
\ifCLASSINFOpdf
\else
\fi

\begin{document}
\title{Few-Example Object Detection \\with Model Communication}

\author{Xuanyi Dong,
        Liang Zheng,
        Fan Ma,
        Yi Yang,
        Deyu Meng
\IEEEcompsocitemizethanks{
\IEEEcompsocthanksitem Xuanyi Dong, Liang Zheng, and Fan Ma are with Centre for Artificial Intelligence, University of Technology Sydney, NSW, Australia. (e-mail: xuanyi.dong@student.uts.edu.au; liangzheng06@gmail.com; fan.ma@student.uts.edu.au)\protect
\IEEEcompsocthanksitem Yi Yang is with Centre for Artificial Intelligence, University of Technology Sydney, NSW, Australia, and State Key Laboratory of Computer Science, Institute of Software, Chinese Academy of Sciences, Beijing, China. (e-mail: yi.yang@uts.edu.au)\protect
\IEEEcompsocthanksitem Deyu Meng (corresponding author) is with School of Mathematics and Statistics and Ministry of Education Key Lab of Intelligent Networks and Network Security, Xi'an Jiaotong University, Shaanxi, P.R. China. (e-mail: dymeng@mail.xjtu.edu.cn)\protect
\IEEEcompsocthanksitem This research was supported by the National Key R\&D Program of China (2018YFB1004300), the Data to Decisions CRC (D2D CRC), the Cooperative Research Centres Programme and the China NSFC projects under contracts 61661166011, 11690011, 61603292, 61721002.
Liang Zheng is partially supported by SIEF STEM+ Business Fellowship Program.
\protect
}
}

\markboth{Accepted to IEEE TRANSACTIONS ON PATTERN ANALYSIS AND MACHINE INTELLIGENCE, 2018. DOI: 10.1109/TPAMI.2018.2844853}
{Shell \MakeLowercase{\textit{et al.}}: Bare Demo of IEEEtran.cls for Computer Society Journals}

\IEEEtitleabstractindextext{
\begin{abstract}
In this paper, we study object detection using a large pool of unlabeled images and only a few labeled images per category, named ``few-example object detection''.
The key challenge consists in generating trustworthy training samples as many as possible from the pool.
Using few training examples as seeds, our method iterates between model training and high-confidence sample selection.
In training, easy samples are generated first and, then the poorly initialized model undergoes improvement.
As the model becomes more discriminative, challenging but reliable samples are selected.
After that, another round of model improvement takes place.
To further improve the precision and recall of the generated training samples, we embed multiple detection models in our framework, which has proven to outperform the single model baseline and the model ensemble method.
Experiments on PASCAL VOC'07, MS COCO'14, and ILSVRC'13 indicate that by using as few as three or four samples selected for each category, our method produces very competitive results when compared to the state-of-the-art weakly-supervised approaches using a large number of image-level labels.
\end{abstract}

}

\maketitle

\IEEEdisplaynontitleabstractindextext

\IEEEpeerreviewmaketitle

\IEEEraisesectionheading{\section{Introduction}\label{sec:introduction}}

\begin{table*}[t]
\setlength{\tabcolsep}{9.5pt}
\centering
\caption{
Comparison of different supervision information used in weakly (semi-) supervised and few-example object detection algorithms. [I] and [V] denote the image and video dataset, respectively. Strong supervision provides the fully annotated images or videos; weak supervision only provides image-level or video-level labels. Data without supervision does not provide any annotation.
Our method consumes negligible annotation efforts compared to others.
}
\begin{supertabular}{| l | l  l  l | l |}\hline
Methods                                    & Data with Strong Supervision  & Data with Weak Supervision   & Data without Supervision   & Test Dataset                   \\\hline
\multirow{2}{*}{\cite{liang2015towards}}   & [I] Flickr; PASCAL VOC        & \multirow{2}{*}{[V] YouTube} & \multirow{2}{*}{-}         & \multirow{2}{*}{PASCAL VOC}    \\
										   & [I] ILSVRC2013-DET            &                              &                            &                                \\\hline
\multirow{2}{*}{\cite{singh2016track}}     & \multirow{2}{*}{-}            & [I] PASCAL VOC               & \multirow{2}{*}{-}         & \multirow{2}{*}{PASCAL VOC}    \\
                                           &                               & [V] YouTube-Object           &                            &                                \\\hline
\multirow{2}{*}{\cite{misra2015watch}}     & [I] Flickr                    & \multirow{2}{*}{-}           & [V] Part of VIRAT          & VIRAT                          \\
                                           & [V] Part of VIRAT and KITTI   &                              & [V] Part of KITTI          & KITTI                          \\\hline
\multirow{2}{*}{\cite{Rochan_2015_CVPR,Tang_2016_CVPR,hoffman2014lsda}}
                                           & [I] ILSVRC2014                & \multirow{2}{*}{-}           & [I] PASCAL VOC 2007        & PASCAL VOC                     \\
                                           & [V] Part of YouTube-Object    &                              & [V] Part of YouTube-Object & YouTube-Object                 \\\hline
\cite{Bilen16,li2016weakly,oquab2015object,tang2017multiple}
                                           & -                             & [I] PASCAL VOC               &      -                     & PASCAL VOC                     \\\hline
\cite{felzenszwalb2010object,wang2014weakly,li2016weakly,diba2016weakly}
                                           & -                             & [I] ILSVRC2013               &      -                     & ILSVRC2013                     \\\hline
\multirow{2}{*}{\cite{wang2015model}}      & [I] 10-200 images per class   & \multirow{2}{*}{[I] SUN}     & \multirow{2}{*}{[I] SUN}   & \multirow{2}{*}{SUN}           \\
										   & on SUN; PASCAL VOC            &                              &                            &                                \\\hline
\multirow{2}{*}{\bf Ours}                  & [I] 3-4 images per class      & \multirow{2}{*}{-}           & \multirow{2}{*}{[I] PASCAL VOC}& \multirow{2}{*}{PASCAL VOC}\\
                                           & on PASCAL VOC                 &                              &                            & \\\hline
\end{supertabular}
\label{table:comparision_annotation}
\end{table*}

\IEEEPARstart{T}{his} paper considers the problem of generic object detection with very few training examples (bounding boxes) per class, named ``few-example object detection (FEOD)''.
Existing works on supervised/semi-supervised/weakly-supervised object detection usually assume much more annotations than this paper. Specifically, we annotate all the bounding boxes in such a number of images that each class will only have 3-4 annotated examples. This task is extremely challenging due to the scarcity of labels which leads to the difficulty in label propagation and model training.

We provide a brief discussion on the relationship between FEOD and other types of supervisions, excluding the methods using strong labels~\cite{dai2016r,girshick2015fast,liu2015ssd,redmon2015you,ren2017faster,gao2013infrared}.
First, strictly speaking, FEOD is a semi-supervised task. But to the best of our knowledge, most works on semi-supervised object detection (SSOD) assume around 50\% of all the labeled bounding boxes \cite{hoffman2014lsda,Rochan_2015_CVPR,Tang_2016_CVPR}.
These methods assume that some classes have strong bounding box labels, while other classes have weak image-level labels \cite{hoffman2014lsda,Rochan_2015_CVPR,Tang_2016_CVPR,wang2018collaborative}. Therefore, FEOD is distinctive from SSOD in terms of the small number of required labels. Second, weakly supervised object detection (WSOD) usually relies on image-level labels \cite{dai2016r,girshick2015fast,liu2015ssd,redmon2015you,ren2017faster}, a type of supervision that is distinct from bounding box level labels as used in FEOD.
An advantage of FEOD over WSOD is that the labeling effort of FEOD is much smaller.
In this paper, we mainly compare our method with the state-of-the-art WSOD works. The third category leverages tracking to mine labels from videos \cite{misra2015watch,singh2016track}.
Usually, these methods focus on moving objects, e.g., car and bicycle, which can be tracked based on their motions along time.
So a potential problem of methods in this category is its effectiveness on stationary objects, e.g., table and sofa, for which tracking may be infeasible.
Table~\ref{table:comparision_annotation} presents a brief summary of the types of supervision used in previous related object detection methods.

Therefore, comparing with the other types of supervision listed in Table~\ref{table:comparision_annotation}, the advantage of FEOD is mainly four-fold.
First, FEOD reduces the labeling effort by using only several annotated bounding boxes per class.
Second, FEOD provides robust supervision to rare classes such as Dugong, where only a few training images can be found. For these classes, image-level supervision on the limited number of images is always not enough to train a good detector.
Third, FEOD can deal with stationary objects, so that it has a larger application scope.
Fourth, FEOD provides accurate annotations to crowded objects, while models trained with image-level labels usually perform poorly on the crowded objects, such as people and bottle.
In comparison, using a few images with bounding box annotations, FEOD can enhance the detector to be robust to such crowded objects.
This can be seen in our experiments. Table~\ref{table:map_voc2007test} evidently shows that the best weakly-supervised algorithm can only achieve 24.7\% mAP on the class of person, but we achieve 40.1\% mAP.
In this paper, we explore the setting in which there is no motion information and no image-level supervision, and there are only several instance-level annotations.
Under this setting, FEOD is extremely challenging due to the lack of labels.
Addressing this challenging yet interesting task is the focus of this paper.

To be specific, the major challenges are:
(1) generating \emph{reliable} pseudo-annotated samples (high precision),
and (2) finding \emph{possibly many} newly annotated samples (high recall).
Specifically, on the one hand, the training samples should be generated with high confidence, i.e., a high precision to guarantee sound guidance for detector training in the following process.
On the other hand, since more training samples benefit a more discriminative detector, we speculate that the generated training samples should have high recall to provide sufficient knowledge for detector amelioration.
A trade-off clearly exists between the precision and recall requirements.

In this paper, two seamlessly integrated solutions, self-paced learning and multi-modal learning, are used to achieve high precision and recall during training sample generation.
In a nutshell, during the training iterations, the selected training images go from ``easy'' (with relatively high confidence) to ``hard'', and the object detector is gradually promoted.
First, a self-paced learning (SPL) framework, in its optimization process, selects ``easy'' training samples and avoids noisy instances.
Second, we embed multi-modal learning in the SPL. Multiple detection models are incorporated in the learning process. Learning from multiple models accomplishes two goals.
(1) It helps alleviate the local minimum issue of the model training,
and (2) it improves the precision and recall of generated training sample due to knowledge compensation between multiple models.
Note that, since the multiple detection models are jointly optimized, our experiments show that multi-modal learning is far superior to model ensembles.
In addition, prior knowledge, i.e., confidence filtration and non-maximum suppression,
can be injected into this learning scheme to further improve the quality of selected training samples.

The major points of this work are outlined below:

\begin{itemize}
\item We address object detection from a new perspective: using very few annotated bounding boxes per class. We propose to alternate between detector improvement and reliable sample generation, thereby gradually obtaining a stable yet robust detector.

\item To ameliorate the trade-off between precision and recall in training sample generation, we embed multiple detection models in a unified learning scheme. In this manner, our method fully leverages the mutual benefit between multiple features and the corresponding multiple detectors.

\item Our proposed algorithm is capable of producing competitive accuracy to state-of-the-art WSOD algorithms, which require much more labeling efforts.
\end{itemize}

\section{Related Work}\label{sec:relate_work}

\subsection{Supervised Object Detection}

Object detection methods based on convolution neural networks (CNNs) can be divided into two types: proposal-based and proposal-free~\cite{girshick2014rich,girshick2015fast,ren2017faster,liu2015ssd,redmon2015you}.
The road-map of proposal-based methods starts from R-CNN~\cite{girshick2014rich}
and is improved by SPP-Net~\cite{he2014spatial} and Fast R-CNN~\cite{girshick2015fast} in terms of accuracy and speed.
Later, Faster R-CNN~\cite{ren2017faster} uses the region proposal network to quickly generate object regions, achieving a high recall compared to previous methods~\cite{uijlings2013selective,zitnick2014edge}.
Many methods directly predict bounding boxes without generating region proposals~\cite{liu2015ssd,redmon2015you}.
For example, YOLO~\cite{redmon2015you} uses the whole feature map from the last convolution layer.
SSD~\cite{liu2015ssd} makes improvements by leveraging default boxes of different aspect ratios for multiple feature maps.
These supervised methods require strong supervision, which is relatively expensive to obtain in practice.

\subsection{Semi-supervised Object Detection}
Current SSOD literature usually uses both the image-level labels and some of the bounding box labels.
For example, Yang~\etal~\cite{yang2013semi} design methods to learn video-specific features to boost detection performance.
Liang~\etal~\cite{liang2015towards} propose an elegant method by integrating prior knowledge modeling, exemplar learning and video context learning for the SSOD task.
They utilize around 350k images with bounding box annotations to provide a good initialization for fine-tuning the detection model on PASCAL VOC.
Besides, they use a negative dataset (without the 20 classes on VOC) as well as around 20k labeled videos. In comparison, our algorithm only requires 3-4 bounding boxes of the target classes, e.g., 20 classes on PASCAL VOC, and do not use any outsider dataset.
Misra~\etal~\cite{misra2015watch} start training with some instance-level annotations and iteratively learn more instances by fusing detection and tracking information.
In \cite{singh2016track}, discriminative visual regions are assigned with pseudo-labels by matching and retrieving technique.
Compared with them, we do not need any extra supervised auxiliary knowledge and the required amount of given annotations is kept at a extremely low level.

\subsection{Weakly Supervised Object Detection}
The WSOD setting is to utilize the image-level label of each image to train object detectors.
Some works employ off-the-shelf CNN models \cite{song2014learning,song2014weakly,wang2014weakly}.
Others design new CNN architectures to obtain object information from the classification loss and leverage this classification model to derive object detectors~\cite{Bilen16,diba2016weakly,li2016weakly,oquab2015object}.
For example, Bilen~\etal~\cite{Bilen16} propose a weakly supervised detection network using selective search (SS)\cite{uijlings2013selective} to generate proposals and train image-level classification based on regional features.
Li~\etal~\cite{li2016weakly} train an image-level classifier to adapt detection results through a mask-out strategy and MIL.
Tang~\etal~\cite{tang2017multiple} integrate a multiple instance detection network and multi-stage instance classifiers in a single network, in which the results of one stage can be used as supervision for the next stage.
Ge~\etal~\cite{GeYY18} propose a weakly supervised curriculum pipeline to jointly optimize recognition, detection, and segmentation, so that multi-task learning enhances the detection performance.
The aforementioned methods depart from our method in that image-level labels are used, which are still expensive to collect when compared with our scheme.

\subsection{Object Detection from Few Examples}
A limited number of previous works can be classified into our settings. Wang~\etal~\cite{wang2015model} propose to generate a large number of object detectors from few samples by model recommendation.
However, they use 10-100 training samples per class, and their initial detectors are required to be trained on other large-scale detection datasets.
Compared to previous methods~\cite{levi2004learning,wang2015model,liang2015towards}, our approach only requires 2-4 examples per class without any extra training datasets.

Here we also briefly introduce and contrast few-shot learning and semi-supervised learning with the few-example learning setting.
On the one hand, few-shot learning~\cite{vinyals2016matching,santoro2016meta,fei2006one,xu2016few} aims to learn a model based on a few training examples without unlabeled data.
In contrast, learning from few samples~\cite{levi2004learning,wang2015model} usually learns an initial model based on the few labeled data, and then progressively ameliorate the initial model on unlabeled data.
An important difference between few-example and few-shot learning is whether to use the unlabeled data.
On the other hand, semi-supervised learning~\cite{yang2013semi,liang2015towards} also leverages a portion of the annotations, which is similar to few-shot learning and few-example learning.
However, semi-supervised learning can use a relatively large number of annotations (e.g., 50\% of the full annotations), which is different from few-example learning and few-shot learning.
We also note that semi-supervised learning can also use only a few annotations. In this scenario, few-example learning is a special case of semi-supervised learning.

\subsection{Webly Supervised Learning for Object Detection}
It can also reduce the annotation cost by leveraging web data.
Chen~\etal~\cite{chen2015webly} propose a two-step approach to initialize the CNN models from easy sample first, and then adapt it to more realistic images.
Divvala~\etal~\cite{divvala2014learning} propose a fully-automated approach for learning extensive models for a wide range of variations via webly supervised learning, while their system requires lots of collection and training time.
Besides, the algorithm can not obtain a good detection model even with 10 million automatically annotated images.
Co-localization algorithms~\cite{tang2014co} localize the objects of the same class across a set of distinct images. They usually leverage the Internet images and are also able to detection objects, but require a strong prior that the image set contains objects with the same class.
Some researchers \cite{rubinstein2013unsupervised,cho2015unsupervised} propose an unsupervised algorithm to discover the common objects from large image collections via the Internet search.
They usually assume the clean labels, but for most object classes, this assumption is unrealistic in real-world settings.

\subsection{Zero-shot Object Detection}
Zero-shot object detection (ZSD)~\cite{rahman2018zero,bansal2018zero,zhu2018zero} aims to locate object instances belonging to novel categories without any training examples. Rahman~\etal~\cite{rahman2018zero} propose a deep network to model the interplay between visual and semantic domain information jointly. Bansal~\etal~\cite{bansal2018zero} adapt visual-semantic embeddings for ZSD, and provide novel splits and baseline experiments on MSCOCO and Visual Genome~\cite{krishna2017visual}.
ZDS is a very challenging task and has many potential research possibilities.
The focus of this paper is not on detecting the new categories of objects like ZDS, while on extracting detectors from extremely few training samples for each class of objects. Thus their purposes are different.

\subsection{Model Ensemble}
Ensemble methods are widely used.
Dai~\etal~\cite{dai2007detector} ensemble multiple part detectors to form sub-structure detectors, which further constitute the final object detector.
The algorithm of \cite{malisiewicz2011ensemble} is based on the linear SVM classifier, which is limited to using the off-the-shelf features.
Yang~\etal~\cite{yang2013feature} use a low rank model to ensemble knowledge learned from different tasks.
Zheng~\etal~\cite{zheng2017discriminatively} fuse the verification and classification models.
Bilen~\etal~\cite{Bilen16} averagely fuse three detection models with different architectures.
Ma~\etal~\cite{ma2017many} suggest assigning different weights of negative examples could improve the detection performance.
Many previous detection methods~\cite{dai2007detector,malisiewicz2011ensemble,Bilen16} employ model ensemble as post-processing.
However, without considering the multiple models in training, these methods may not fully utilize the complementary nature of different detection models. In this paper, we jointly optimize multiple detection models during training to further improve each model.

\subsection{Progressive Paradigm}

Our method adapts a progressive strategy to iteratively optimize the multiple detection models, which is related to curriculum learning~\cite{bengio2009curriculum} and self-paced learning~\cite{kumar2010self}.
Bengio~\etal~\cite{bengio2009curriculum} first propose a learning paradigm in which organizing the examples in a meaningful order significantly improves the performance.
Kumar~\etal~\cite{kumar2010self} propose to determine the training sample order by how easy they are.
Many other researchers~\cite{jiang2014self,jiang2015self,ma2017spacotrain,lin2018active,meng2017SPLInsight} propose more theoretically analysis on this progressive paradigm.
Some researches also apply the similar idea of the progressive paradigm~\cite{yan2016image,dong2017dual,zheng2018sift}.
For example, Wei~\etal~\cite{wei2017stc} propose a simple to complex framework that learns to segment with image-level labels.
Liang~\etal~\cite{liang2017learning} leverage a iterative framework to learn segmentation from YouTube videos.
Our algorithm extends this progressive strategy into multiple model ensemble.
Consequently, we obtain a significantly improvement in object detection from few examples.

\begin{figure*}[!t]
\centering\includegraphics[width=\textwidth]{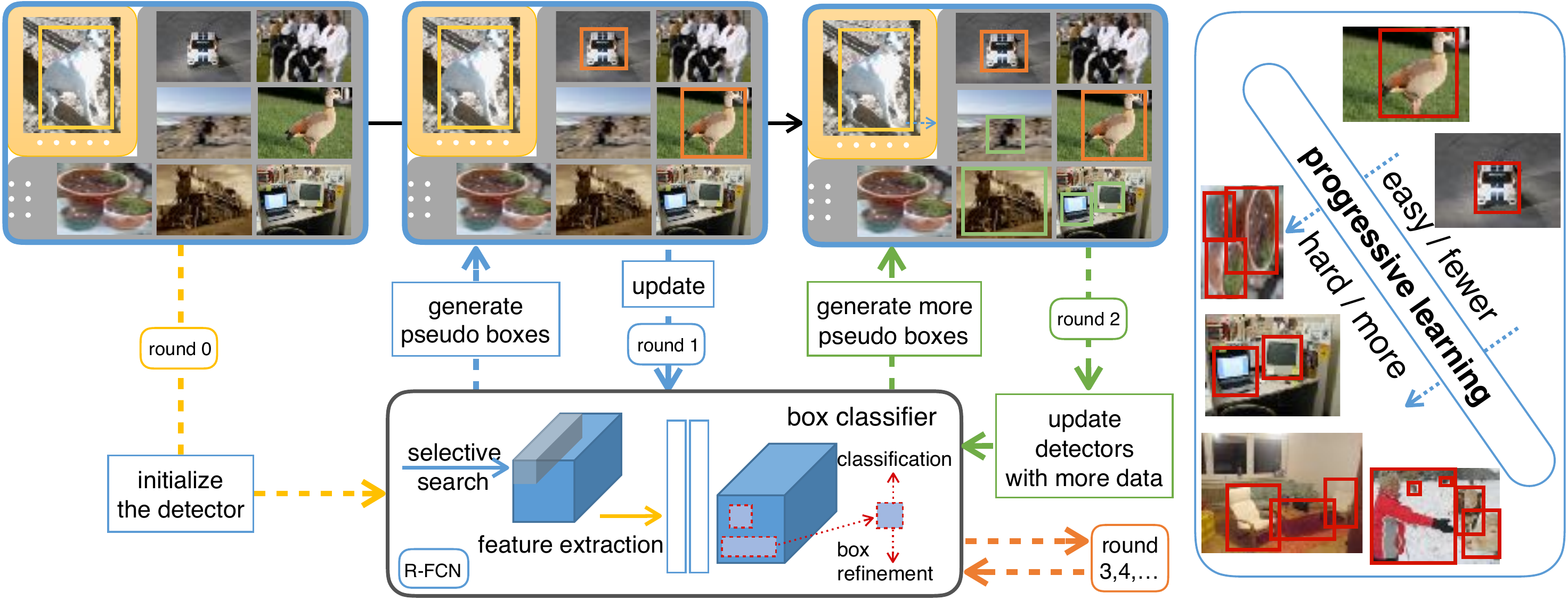}
\caption{A simplified version of MSPLD without multi-modal learning.
The blue boxes in the top row contain the training images
where the few labeled and the many unlabeled images are in the gray and yellow areas, respectively.
The gray solid box represents our detector, R-FCN.
We train the detector using the few annotated images.
The detector generates reliable pseudo instance-level labels
and then gets improved with these pseudo-labeled bounding boxes, as shown round 1.
In the following rounds (iterations), the improved detector can generate larger numbers of reliable pseudo-labels that further update the detector.
When the label generation and detector updating steps work iteratively,
more pseudo boxes are obtained from ``easy'' to ``hard'', and the detector becomes more robust.
}
\label{fig:basic_framework}
\end{figure*}

\section{The Proposed Method}

As our framework combines self-paced learning and multi-modal learning, we call it Multi-modal Self-Paced Learning for Detection (MSPLD).
We first introduce some basic notations in Sec.~\ref{sec:prelim}, and demonstrate the detailed formulation of our MSPLD in Sec.~\ref{sec:mspld}.
Then, we describe the optimization method in Sec.~\ref{sec:opt}.
Lastly, we show the whole algorithm description in Sec.~\ref{sec:alg}.

\subsection{Preliminaries}\label{sec:prelim}
We choose Fast R-CNN~\cite{girshick2015fast} and R-FCN~\cite{dai2016r} as the basic detectors.
Both networks achieve the state-of-the-art performance when provided with strong supervisions.
The Fast R-CNN network uses the Region-of-Interest (RoI) pooling layer and multi-task loss to improve the efficiency and effectiveness.
The R-FCN optimizes the Fast R-CNN with the position-sensitive score maps, and all the computations are shared over the entire image instead of being split for each proposal.
Each detector has a different architecture and thus reflects different,
but complementary, intrinsic characteristics of the underlying samples.
As for the region proposal, we use unsupervised methods because they do not require human annotations and are applicable to handling the situation of few annotations in our setting, such as SS~\cite{uijlings2013selective} and edge box~\cite{zitnick2014edge}.
We denote the proposal generation as function $B$, which takes an image $I$ as input.
For simplification, we denote the detector (Fast R-CNN and R-FCN) as function $F$.
Therefore, the generation of region proposals can be formalized as:
{
\begin{align}
rectangle = (up,left,bottom,right),
\end{align}
\begin{align}
B(I) = \{ rectangle_i | 1 \leq i \leq n \},
\end{align}
}\noindent where each proposal is a rectangle in the image and $(up,left)$ and $(bottom,right)$ represent the coordinates of the upper left corner and the bottom right corner of this rectangle.
The generated proposals are likely to be the true objects.
We then have
{
\begin{align}
\label{eq:cnn_out}
F(I,B(I)) = \{ (rectangle, score)_{(i,j)} | 1 \leq i \leq n, 1 \leq j \leq C \},
\end{align}
}\noindent where $C$ is the number of object classes, $score$ represents the confidence score for the corresponding proposal.

\subsection{The MSPLD Model}\label{sec:mspld}
Suppose we have $l$ labeled images in which all the object bounding boxes are annotated.
Note that, when we randomly annotate approximately four images for each class, an image may contain several objects, and we annotate all the object bounding boxes.
We denote the labeled images as $y_i \subset [{\mathbb R}^4, C], i = 1,...l$.
We also have $u$ unlabeled images $y_i^u \subset [{\mathbb R}^4, C], i = 1,...u$.
The unlabeled bounding boxes will be assigned labels, or discarded during each training iteration.
We also assume there are $m$ detection models.
In technical terms, our method integrates multi-modal learning into the SPL framework.
Our model can be formulated as Eq.~\eqref{eq:more_spld}, Eq.~\eqref{eq:more_spld_st_v1}, Eq.~\eqref{eq:more_spld_st_v2} and Eq.~\eqref{eq:more_spld_st_v3}.
{
\begin{align}
\label{eq:more_spld}
E(w^{j},v^j_{i,c},{y^u}^{j}_{i};\lambda,\Psi)= \sum_{j=1}^{m} \sum_{i=1}^{l} L^{j}_{s}(y_{i},I_i,B(I_i),w^j) \nonumber \\
+ \sum_{j=1}^{m} \sum_{i=1}^{u} \sum_{c=1}^{C} v_{i,c}^j L^{j}_{c}({y^{u}}^{j}_{i},I_i,B(I_i),w^j) \nonumber \\
- \sum_{j=1}^{m} \sum_{i=1}^{u} \sum_{c=1}^{C} \lambda^{j}_{c} v^{j}_{i,c}
- \sum_{j_{1}=1}^{m} \sum_{j_{2}=j_{1}+1}^{m} \gamma^{j_1,j_2}(V^{j_1})^{T}{V^{j_2}}  \\
\label{eq:more_spld_st_v1}
s.t.~ \sum_{c=1}^{C} v^{j}_{i,c} \leq 1 ~~~for~~~1 \leq j \leq m ~~\&~~ 1 \leq i \leq u,  \\
\label{eq:more_spld_st_v2}
      v^{j}_{i,c} \in \{0,1\}~~\&~~ v \in \Psi_{v},                                  \\
\label{eq:more_spld_st_v3}
      {y^{u}}^{j}_{i} \subset F^{*}(I_{i},B(I_{i}),w) ~~and~~ {y^{u}}^{j}_{i} \in \Psi_{y} ~~for~~1 \leq i \leq u,
\end{align}
}\noindent where $w^{j}$ denotes the parameters of the $j^{th}$ basic detector.
$v^{j}_{i,c}$ encodes whether the bounding boxes in the $i^{th}$ image are determined as the $c^{th}$ class to train the $j^{th}$ model.
Thus, $v^{j}_{i,c}$ can only be 0 or 1.
${y^u}^{j}_{i}$ is the generated pseudo bounding boxes for the unlabeled images from the $j^{th}$ detector.
$i,j,c$ are the indexes of images, models, and classes, respectively.
$V^{j}$ is a $u \times C$ matrix and denotes all the $v^{j}_{i,c}$ for the $j^{th}$ detection model.
$\lambda$ is the parameter for the SPL regularization term,
which enables the possibly selection of high confidence images during optimization.
$\gamma$ is the parameter for the multi-modal regularization term.
Note that an inner product regularization term $(V^{i})^{T}V^j$ has been imposed on each pair of selection weights $V^{i}$ and $V^{j}$.
This term delivers the basic assumption that different detection models share common knowledge of pseudo-annotation confidence for images, i.e., an unlabeled image is labeled correctly or incorrectly simultaneously for both models.
This term thus encodes the relationship between multiple models.
It uncovers the shared information and leverages the mutual benefits among all the models.

\begin{figure}[t]
\centering\includegraphics[width=\linewidth]{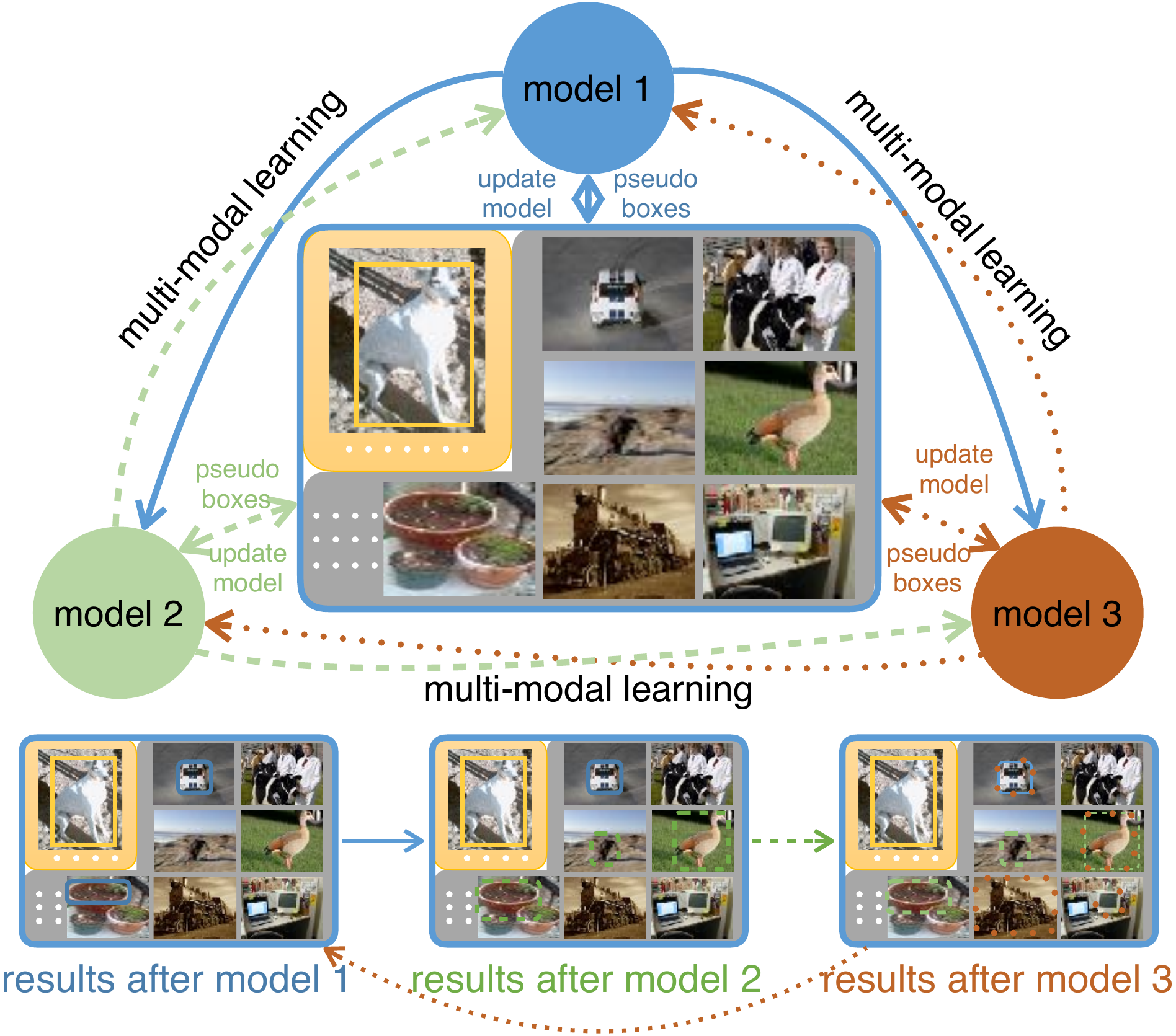}
\caption{
The working flow of MSPLD when multi-modal learning is integrated with Figure~\ref{fig:basic_framework}.
An example with three models is shown.
The three discs with different colors indicate the basic detectors.
The images in the middle are the training data. The three detectors complement each other in validating the selected training samples.
For example, as shown in the bottom row, the 1$^{st}$ model only detects two objects and misalignment exists with the detected plant.
The 2$^{nd}$ model detects three other objects.
When considering the detections of the 1$^{st}$ model, the misaligned plant is corrected, and the car with the blue box is also used to train the 2$^{nd}$ model.
So more training data with reliable labels are used to improve the performance of model 2.
Similarly, the 3$^{rd}$ model obtains more pseudo boxes and gets updated in turn. The whole procedure iterates until convergence.
}
\label{fig:co_training}
\end{figure}

In Eq.~\eqref{eq:more_spld}, $L_s$ represents the original multi-task loss of the supervised object detection \cite{girshick2015fast,girshick2014rich,ren2017faster}.
The loss function for the unlabeled images $L_{c}$ is defined as
{
\begin{align}
L_{c} =
  \begin{cases}
     L_{s}   & \mbox{if } the~c^{th}~class~appears~in~y_i \\
     \infty  & \mbox{if } otherwise
  \end{cases}.
\end{align}
}\noindent Given the constraints in Eq.~\eqref{eq:more_spld_st_v1} and Eq.~\eqref{eq:more_spld_st_v2}, it is guaranteed that $L_{s} = \sum_{c=1}^{C} v_{i,c}^j L_{c}$ if the $i^{th}$ image is selected as the training data by the $j^{th}$ detection model.
As the distribution of the confidence/loss can be different for different classes,
this class-specific loss function helps the selected images cover as many classes as possible.
$F^{*}$ indicates the fused results from multiple models, which contains $n\times C$ bounding boxes and, thus, has too many noisy objects.
We use some empirical procedures to select the faithful pseudo-objects,
and incorporate prior knowledges into a curriculum regime $y^{u} \in \Psi_{y}$.
Similar to $\Psi_{y}$, some specially designed processes for discarding the unreliable images is denoted as $v \in \Psi_{v}$.
The detailed steps of $\Psi_{y}$ and $\Psi_{v}$ will be discussed in the next section.

\subsection{Optimization}\label{sec:opt}

We adopt the alternative optimization strategy~(AOS) to solve Eq.~\eqref{eq:more_spld}.
The parameters are iteratively updated by the sequence $y^{u1}, v^{1}, w^{1},... v^{j}, y^{uj}, w^{j}, y^{u1}, v^{1}, w^{1}...$~{until there are no more available unlabeled data or the maximum iteration number is reached.}
In this section, we show how to solve each parameter as follows.

{\bf Update $v^{j}$}: This step aims to update the training pool of the $j^{th}$ detection model.
We can calculate the derivative of Eq.~\eqref{eq:more_spld} with respect to $v^{j}_{c}$ as:
{
\begin{align}
\label{problem_v}
\frac{\partial{E}}{\partial{v^{j}_{i,c}}} = L_{c}({y^{u}}^{j}_{i},I_i,B(I_i),w^j)
                      - \lambda^{j}_{c} - \sum_{k=1;k\neq j}^{m}\gamma^{j,k} v^{k}_{i,c} .
\end{align}
}\noindent Then the closed-form solution is
{
\begin{align}
\label{eq:update_v}
v^{j}_{i,c} =
  \begin{cases}
     1 & \mbox{if } L^{j}_{i,c} < \lambda^{j}_{c} + \sum_{k=1;k\neq j}^{m}\gamma^{j,k} v^{k}_{i,c} \\
     0 & \mbox{if } L^{j}_{i,c} \geq \lambda^{j}_{c} + \sum_{k=1;k\neq j}^{m}\gamma^{j,k} v^{k}_{i,c}
  \end{cases} ,
\end{align}
}\noindent for the unlabeled images.
Due to the limitation of $\sum_{c=1}^{C} v^{j}_{i,c} \leq 1$,
if there are multiple $v^{j}_{i,c}=1$ for the same $(i,j)$ indicating the same image,
we only choose the one with the lowest corresponding loss value $L^{j}_{i,c}$.
The item $\gamma$ and $v^{k}_{i,c}$ uncover the shared information.
Because if $v^{k}_{i,c}=1$ (indicate the i$^{th}$ image is selected by the k$^{th}$ model) the threshold in Eq.~\eqref{eq:update_v} will become higher, and this image will become easier to be selected by the current detector.

\begin{algorithm}[!t]
\caption{AOS for Solving MSPLD}
\label{alg:more_spld}
\begin{algorithmic}[1]

\Require ${\mathbb L}=\{(x_i^l,y_i)\}$ and ${\mathbb U}=\{(x_i^u)\}$ \\
~~~~~~$m$ basic detectors with parameters $W$ \\
~~~~~~$\lambda$, $\gamma$, $\Psi_v$, $\Psi_y$ and max iteration

\State initialize $W$ trained by ${\mathbb L}$
\State initialize $V_{j}=O$ for $1\leq j\leq m$

\For{$iter=1$; $iter\leq max$; iter++}

\For{$j=1$; $j\leq m$; j++}

      \State Clean up the unlabeled data via curriculum $\Psi_{v}$

      \State Generate the pseudo labels $y_i^{u}$ via Eq.~\eqref{eq:update_y}

      \State Compute loss $L^j_c$ by $j^{th}$ detector~\cite{girshick2015fast,dai2016r}

      \State Update $V_j$ according to Eq.~\eqref{eq:update_v}
      \State Update ${y^{u}}^{j}$ and $V_j$ via the prior knowledge

      \State Retrain $w_j$ via training pool $\{(x_i^u,y_i^{uj})\}~\cup~{\mathbb L}$
\EndFor
   \State Update $\lambda$, $\gamma$ to select more images in the next round
\EndFor

\Ensure detectors' parameters $W=\{w_j | 1\leq j\leq m\}$

\end{algorithmic}
\end{algorithm}

{\bf Update $w^{j}$}:
We will train the basic detector of the $j^{th}$ model, given $v$ and $y^{u}$.
The training data is the union set of initial annotated images and the selected images ($v^{j}_{i,c}=1$) with the pseudo boxes $y^{u}$.
Due to the limitation of $\sum_{c=1}^{C} v^{j}_{i,c} \leq 1$ and $v^{j}_{i,c} \in \{0,1\}$, our selected images are unique.
Finally this step can be solved by the standard process, described as~\cite{dai2016r,girshick2015fast}.

{\bf Update ${y^{u}}^{j}$}: Fixing $v$ and $w$, ${y^{u}}^{j}$ should be solved by the following minimization problem:
{
\begin{align}
\label{eq:update_y}
{y^{u}}^{j}_{i} = \arg\min_{{y^{u}}^{j}_{i}} \sum_{j=1}^{m} \sum_{c=1}^{C} v_{i,c}^j L_{c}({y^{u}}^{j}_{i},I_i,B(I_i),w^j) , \nonumber \\
s.t.~~{y^{u}}^{j}_{i} \subset F^{*}(I_{i},B(I_{i}),w)~~for~~1\leq i \leq u
\end{align}
}
\noindent It's almost impossible to directly optimize ${y^{u}}^{j}_{i}$, because ${y^{u}}^{j}_{i} \subset [{\mathbb R}^4, C]$ is a set of bounding boxes.
Hence, we leverage prior knowledges to empirically calculate pseudo boxes ${y^{u}}^{j}_{i}$.
We fuse the results from all detection models and obtain the outputs of $F*$.
Then the post-processes of NMS and thresholding are applied on $F*$ to generate ${y^{u}}^{j}_{i}$.

\subsection{Algorithm Description}\label{sec:alg}

We summarize MSPLD in Algorithm~\ref{alg:more_spld}.
The 7$^{th}$/11$^{th}$ steps are prior constrains to filter unreliable images, corresponding to Eq.~\eqref{eq:more_spld_st_v3}.
The 8$^{th}$ and 12$^{th}$ steps are the solution for updating $y_i^{u}$ and $W$, respectively (see the second and third paragraphs in Sec.~\ref{sec:opt}).
The 9$^{th}$/10$^{th}$ steps are used to update $V$ via the SPL and multi-modal regularization terms.
Later, we illustrate this optimization process in Figure~\ref{fig:basic_framework} and Figure~\ref{fig:co_training}.

Figure~\ref{fig:basic_framework} illustrates a special case of our MSPLD with only one detection model, which means the case of $m$=1 in Eq.~\eqref{eq:more_spld}.
We initialize the detector with few annotated bounding boxes.
In the 1$^{st}$ round, we generate pseudo boxes with high confidences from some of the unlabeled images and retrain the detector by combining the strongly-labeled and the newly-labeled bounding boxes.
In the next round, with the improved detector, we are able to generate more reliable pseudo boxes, such as the green boxes generated in round 2.
Therefore, the process iterates between instance-level label generation and detector updates.
Through these iterations, our approach gradually generates more bounding boxes with reliable labels, from ``easy'' to ``hard'', shown in Figure~\ref{fig:basic_framework}, and we can, therefore, obtain a more robust detector with these newly labeled training data.

Since this method only uses very few training samples per category, a simple self-paced strategy may be trapped by local minimums.
To avoid this problem, we incorporate multi-modal learning into the learning process, which corresponds to the case of $m>1$ in Eq.~\eqref{eq:more_spld}.
In Figure~\ref{fig:co_training}, we observe that the three detection models are complementary to each other.
These different models can communicate with each other by the multi-modal regularization term.
Each detector can communicate with each other by the effect of $\gamma$ and the prior knowledge in Eq.~\eqref{eq:more_spld}.
At the instance level, the current detector may either correct or directly use the previous results.
For example, the green box of the plant is better aligned by the 2$^{nd}$ model compared to the 1$^{st}$ model; the blue box of the car detected by the 1$^{st}$ model is directly used by the 2$^{nd}$ model.
At the image level, the previously selected images will be assigned higher priority in the next round, see Eq.~\eqref{eq:update_v}.
Besides, the probability of the unselected images remains unchanged.

\begin{figure*}[!t]
\subfloat[][$\lambda$ over the training iterations]{\includegraphics[width=\textwidth]{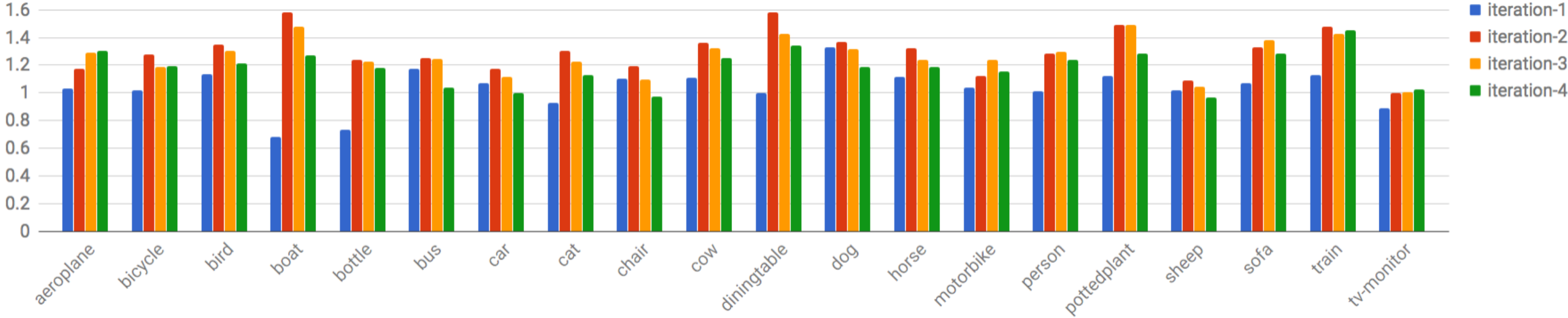}\label{subfig:lambda}}\vfill
\subfloat[][Precision \& Recall]{\includegraphics[width=0.29\textwidth]{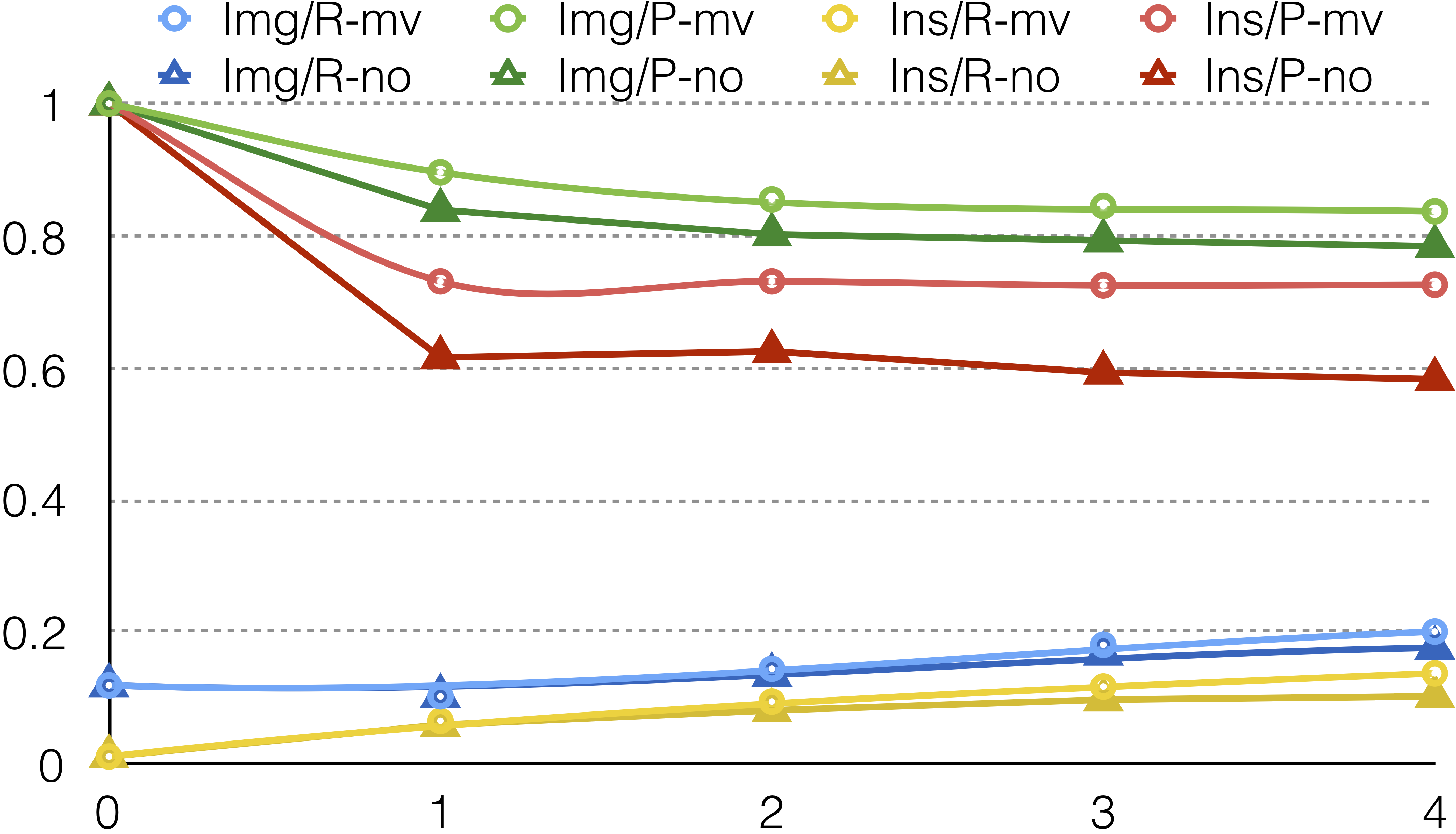}\label{subfig:gt_pdf}}\hfill
\subfloat[][Mean AP]{\includegraphics[width=0.29\textwidth]{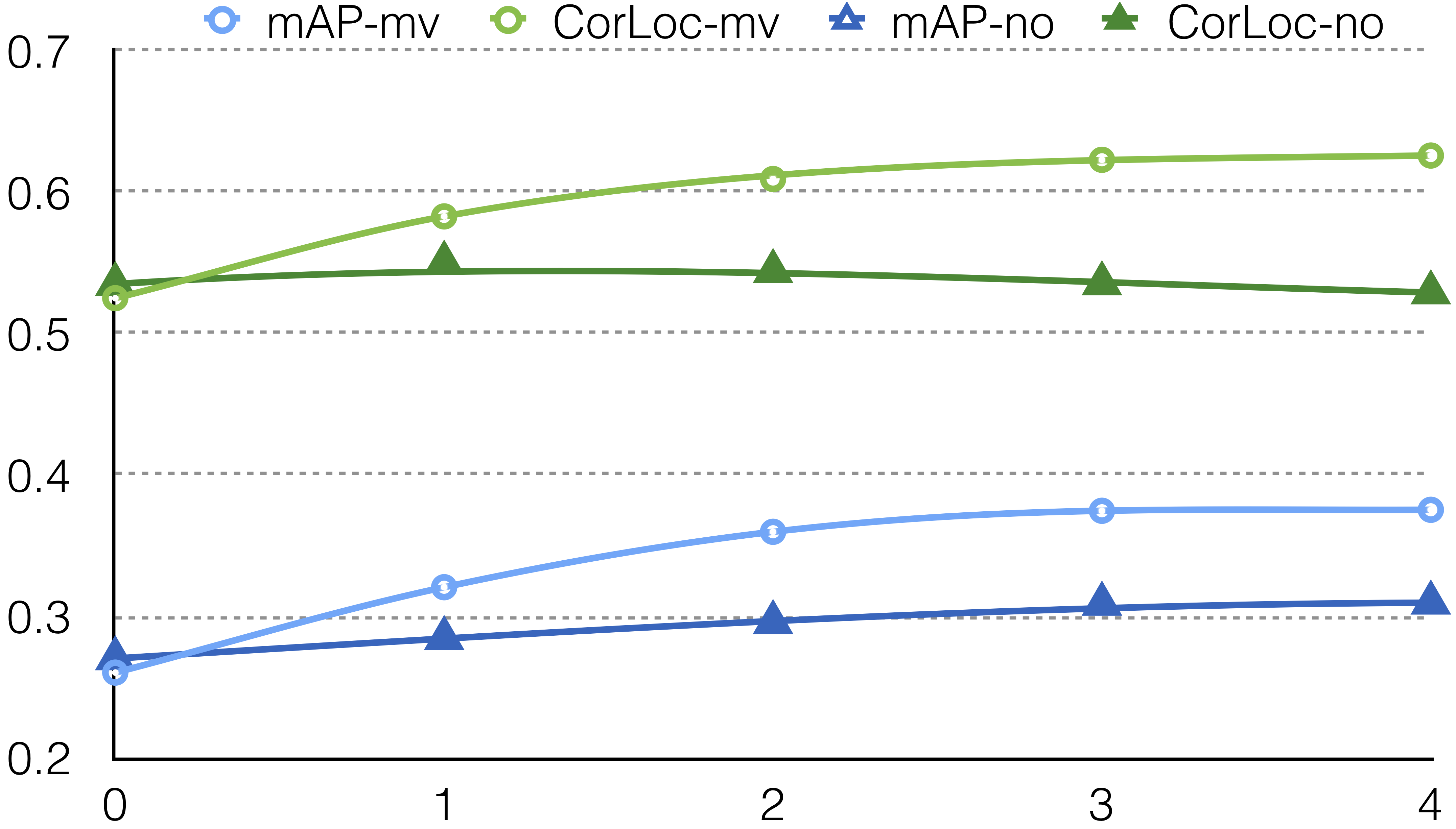}\label{subfig:er_pdf}
  }\hfill
\subfloat[][Number of mined objects and images]{
  \includegraphics[width=0.31\textwidth]{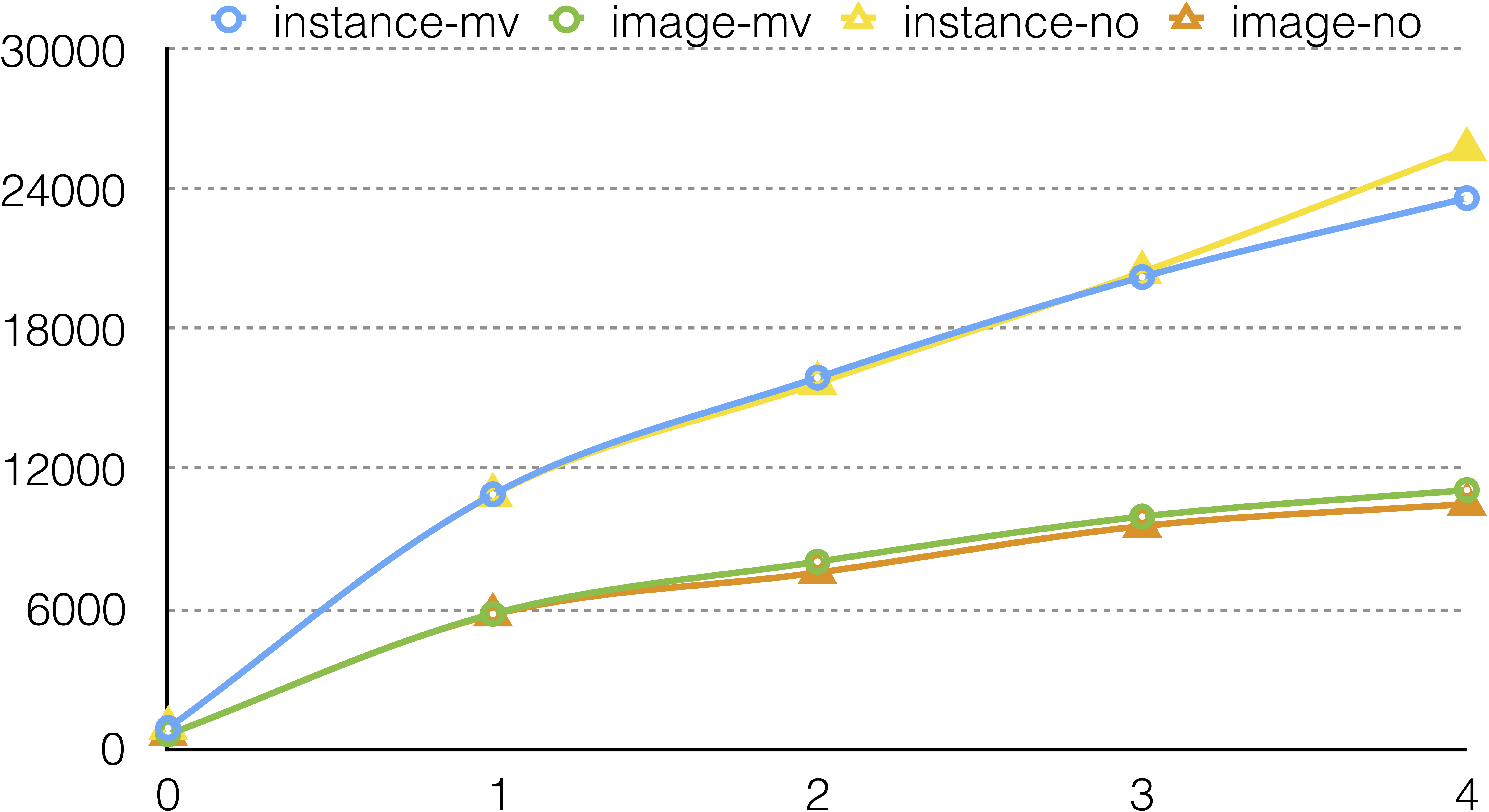}\label{subfig:re_pdf}
  }
\caption{
The change of $\lambda$, precision, recall and mAP for the first four training iterations of MSPLD.
``mv'' and ``no'' denote using and not using multi-modal learning, respectively.
``Img/R'' and ``Ins/R'' indicate the image-level and instance-level recall, respectively.
``Img/P'' and ``Ins/P'' indicate the image-level and instance-level precision, respectively.
}
\label{fig:parameters}
\end{figure*}

The multi-modal mechanism pulls the self-paced baseline out of the local minimum by significantly improving the precision and recall of training objects and images.
In Figure~\ref{fig:parameters}, we show the details of precision/recall using the ResNet-101 model and compare it to the method without multi-modal.
We observe that, as the model iterates, the recall of the training data improves, while the precision decreases, which clearly demonstrates the trade-off between precision and recall.
Meanwhile, the mean average precision (mAP) of object detection keeps increasing and remains stable when precision and recall reach convergence.
Compared with the baseline (no multi-modal),
the precision of images\footnote{``Image-level label'' denotes which objects appear in an image.} (denoted as ``Img/P'')
and instances\footnote{``instance-level label'' denotes (1) the type of the object instance and (2) the instance's location (coordinates) in terms of a rectangular bounding box.} (denoted as ``Ins/P'') is improved by about 6\% and 13\% using multi-modal;
the recall of generated objects and selected images is improved by more than 5\%.
These observations suggest that the multi-modal mechanism obtains a better trade-off between precision and recall.

There are two regularization parameters, $\lambda$ and $\gamma$, in our objective function~Eq.~\eqref{eq:more_spld}. We show how $\lambda$ changes during the training procedures in Figure~\ref{fig:parameters}. As $\lambda$ is related to how many images are used during the training procedure. Therefore, we should use the appropriate parameter $\lambda$ to guarantee the images in the training pool can stably increase over the training iterations.
$\gamma$ is usually fixed as 0.2/(m-1).
More details can be found in experiments.

\begin{figure}[!t]
\centering\includegraphics[width=\linewidth]{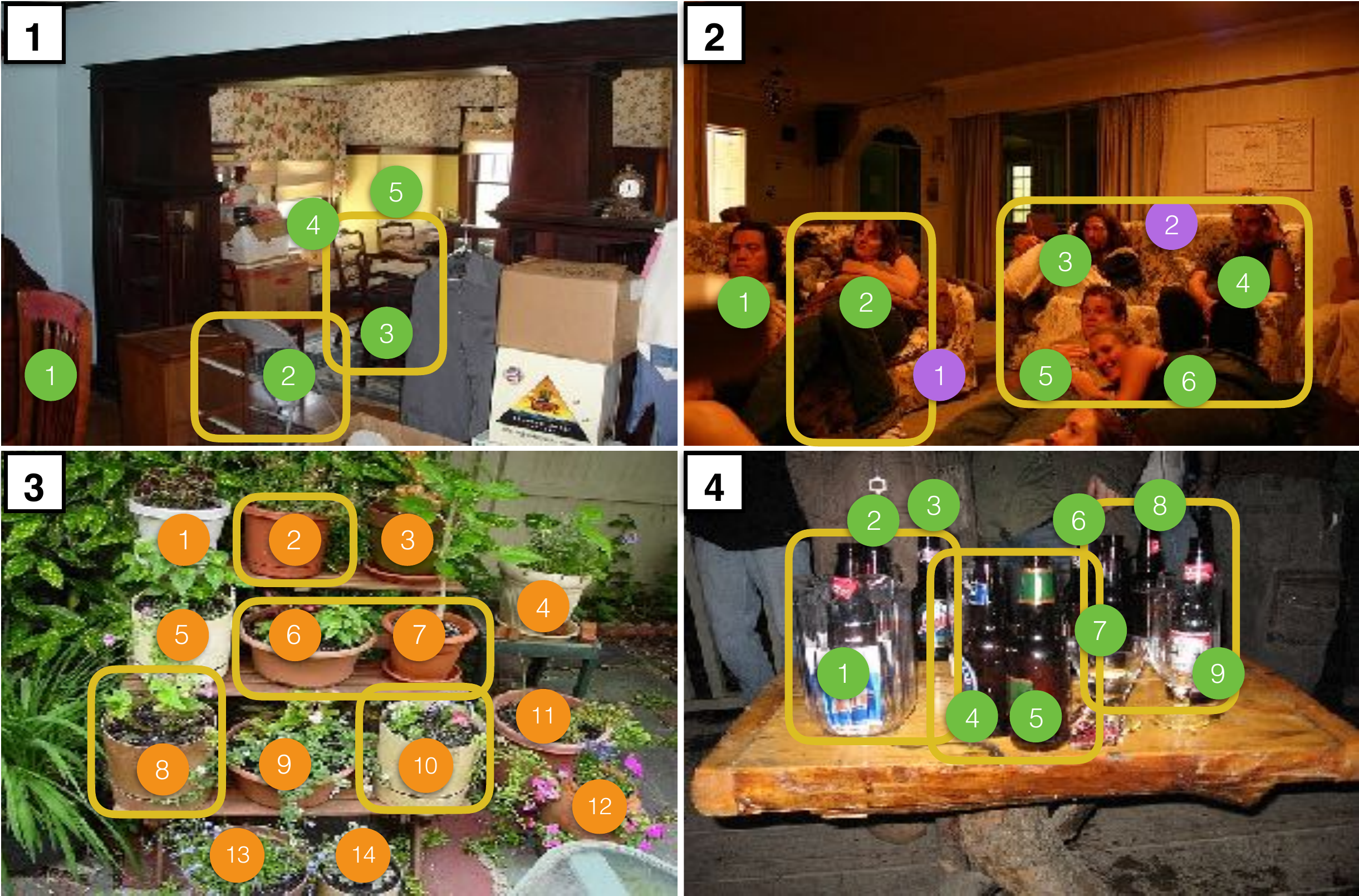}
\caption{
Some poorly located or missed training samples.
The yellow rectangles are the generated labeled boxes, and the discs denote the ground-truth objects.
In image 2, the green and purple circles indicate people and sofa, respectively.
We observe that the sofa is missed due to occlusions and different people are not well separated.
}
\label{fig:filter_image}
\end{figure}

{\bf Injecting prior knowledge.}
In Eq.~\eqref{eq:more_spld_st_v2} and Eq.~\eqref{eq:more_spld_st_v3}, prior knowledge $\Psi_{v}$ and $\Psi_{y}$ are leveraged to filter out some very challenging instances.
For example, as suggested in Figure~\ref{fig:filter_image}, an image could be very complex and it may be challenging to locate the correct bounding box.
Therefore, we empirically design a method to estimate the number of boxes for each class in an image.
Specifically, we apply a non-maximum suppression (NMS) on the output of $F^{*}$ for each class,
and then use a confidence threshold of 0.2.
Later, we employ NMS to filter out the nested boxes, which usually occurs when there are multiple overlapping objects.
If there are too many boxes ($\geq 4$) for one specific class or too many classes ($\geq 4$) in the image, this image will be removed.
To generate relatively robust pseudo instance-level labels (Eq.~\eqref{eq:more_spld_st_v3}), a class-specific threshold is applied on the remaining boxes to select the instance-level instances with high confidence.
Additionally, images in which no reliable pseudo objects are found are filtered out.

\begin{table*}[t]
\setlength{\tabcolsep}{2.4pt}
\centering
\caption{
Method comparisons in average precision (AP) on the PASCAL VOC 2007 test set.
$^{*}$ indicates the usage of full image-level labels for training.
Our approach (the last four rows) requires only approximately four strong annotated images per class.
\cite{zhangbridging2016} leverages the SVM classifier to train the object detector via SPL.
``SPL+Fast R-CNN'' is our approach using only one model, i.e., Fast R-CNN with VGG16,
and ``SPL+R-FCN'' denotes R-FCN with ResNet50$^{ohem}$.
``SPL+Ensemble'' ensembles the three models: Fast R-CNN with VGG16, R-FCN with ResNet50$^{ohem}$ and R-FCN with ResNet101.
}
\begin{tabular}{|l|cccccccccccccccccccc|c|}
\hline
Methods  & aero  & bike & bird & boat & botl & bus  & car  & cat  & chair & cow  &
           table & dog  & hors & mbik & pers & plnt & shp  & sofa & train & tv   & mean  \\\hline
Zhang~\etal\cite{zhangbridging2016}$^{*}$
         & 47.4  & 22.3 & 35.3 & 23.2 & 13.0 & 50.4 & 48.0 & 41.8 & 1.8  & 28.9 &
           27.8  & 37.7 & 41.6 & 43.8 & 20.0 & 12.0 & 27.8 & 22.9 & 48.9 & 31.6 & 31.3 \\
Teh \etal\cite{tehattention}$^{*}$
         & 48.8  & 45.9 & 37.4 & 26.9 & 9.2  & 50.7 & 43.4 & 43.6 & 10.6  & 35.9 &
           27.0  & 38.6 & 48.5 & 43.8 & 24.7 & 12.1 & 29.0 & 23.2 & 48.8  & 41.9 & 34.5 \\
Kantorov \etal\cite{kantorov2016}$^{*}$
         & 57.1  & 52.0 & 31.5 & 7.6  & 11.5 & 55.0 & 53.1 & 34.1 & 1.7   & 33.1 &
           49.2  & 42.0 & 47.3 & 56.6 & 15.3 & 12.8 & 24.8 & 48.9 & 44.4  & 47.8 & 36.3 \\
Bilen \etal\cite{Bilen16}$^{*}$
         & 46.4  & 58.3 & 35.5 & 25.9 & 14.0 & 66.7 & 53.0 & 39.2 & 8.9   & 41.8 &
           26.6  & 38.6 & 44.7 & 59.0 & 10.8 & 17.3 & 40.7 & 49.6 & 56.9  & 50.8 & 39.3 \\
Li \etal\cite{li2016weakly}$^{*}$
         & 54.5  & 47.4 & 41.3 & 20.8 & 17.7 & 51.9 & 63.5 & 46.1 & 21.8  & 57.1 &
           22.1  & 34.4 & 50.5 & 61.8 & 16.2 & 29.9 & 40.7 & 15.9 & 55.3  & 40.2 & 39.5 \\
Diba \etal\cite{diba2016weakly}$^{*}$
         & 49.5  & 60.6 & 38.6 & 29.2 & 16.2 & 70.8 & 56.9 & 42.5 & 10.9 & 44.1 &
           29.9  & 42.2 & 47.9 & 64.1 & 13.8 & 23.5 & 45.9 & 54.1 & 60.8 & 54.5 & 42.8 \\
Dong~\etal\cite{dong2017dual}$^{*}$
		 & 62.5  & 54.6 & 44.3 & 12.9 & 12.7 & 63.8 & 60.6 & 25.0 & 5.4  & 48.0 &
           49.3  & 58.7 & 66.6 & 63.5 & 8.5  & 17.3 & 40.7 & 59.4 & 53.9 & 51.4 & 43.0 \\
Tang~\etal\cite{tang2017multiple}$^{*}$
         & 65.5  & 67.2 & 47.2 & 21.6 & 22.1 & 68.0 & 68.5 & 35.9 & 5.7  & 63.1 &
           49.5  & 30.3 & 64.7 & 66.1 & 13.0 & 25.6 & 50.0 & 57.1 & 60.2 & 59.0 & 47.0 \\
Ge~\etal\cite{GeYY18}$^{*}$
         & 64.3  & 68.0 & 56.2 & 36.4 & 23.1 & 68.5 & 67.2 & 64.9 & 7.1  & 54.1 &
           47.0  & 57.0 & 69.3 & 65.4 & 20.8 & 23.2 & 50.7 & 59.6 & 65.2 & 57.0 & \BF{51.2} \\
\hline
SPL+Fast R-CNN
         & 41.4  & 55.9 & 24.5 & 15.7 & 22.4 & 37.3 & 52.4 & 37.9 & 14.3 & 17.5 &
            33.0 & 27.9 & 41.4 & 50.2 & 36.7 & 19.5 & 27.2 & 46.0 & 47.5 & 26.0 & 33.7$\pm$0.5 \\
SPL+R-FCN
         & 25.6  & 34.3 & 26.0 & 15.3 & 22.3 & 39.3 & 48.8 & 30.4 & 18.8 & 17.3 &
            2.2  & 18.6 & 40.9 & 54.8 & 35.4 & 13.5 & 26.6 & 36.1 & 52.1 & 35.8 & 29.9$\pm$1.1 \\
SPL+Ensemble
         & 38.4  & 51.1 & 41.4 & 21.6 & 25.9 & 45.0 & 57.6 & 50.0 & 22.0 & 21.7 &
             7.5 & 23.8 & 47.4 & 56.0 & 43.4 & 22.1 & 31.3 & 46.1 & 57.8 & 42.0 & 37.6$\pm$0.8 \\
\textbf{MSPLD}
         & 46.6  & 55.6 & 37.9 & 26.1 & 27.9 & 46.6 & 57.9 & 58.1 & 24.1 & 37.6 &
           12.8  & 33.1 & 51.4 & 59.7 & 40.1 & 17.5 & 36.1 & 52.0 & 61.4 & 52.1 & 41.7$\pm$0.3 \\
\hline
\end{tabular}
\label{table:map_voc2007test}
\end{table*}

\begin{table*}[t]
\setlength{\tabcolsep}{2.4pt}
\centering
\caption{
Method comparisons in correct localization (CorLoc~\cite{deselaers2012weakly}) on the PASCAL VOC 2007 trainval set.
$^{*}$ indicates the usage of full image-level labels for training.
The models that we use are the same as Table~\ref{table:map_voc2007test}.
}
\begin{tabular}{|l|cccccccccccccccccccc|c|}
\hline
Methods  & aero  & bike & bird & boat & botl & bus  & car  & cat  & chair & cow  &
           table & dog  & hors & mbik & pers & plnt & shp  & sofa & train & tv   & mean  \\\hline
Zhang \cite{zhangbridging2016}$^{*}$
         & 75.7  & 37.9 & 68.3 & 53.2 & 11.9 & 57.1 & 59.6 & 63.7 & 16.4 & 63.9 &
           17.5  & 62.3 & 71.6 & 71.5 & 45.6 & 14.7 & 53.1 & 41.1 & 75.5 & 24.4 & 49.3 \\
Li \etal\cite{li2016weakly}$^{*}$
         & 78.2  & 67.1 & 61.8 & 38.1 & 36.1 & 61.8 & 78.8 & 55.2 & 28.5  & 68.8 &
           18.5  & 49.2 & 64.1 & 73.5 & 21.4 & 47.4 & 64.6 & 22.3 & 60.9  & 52.3 & 52.4  \\
Bilen \etal\cite{Bilen16}$^{*}$
         & 73.1  & 68.7 & 52.4 & 34.3 & 26.6 & 66.1 & 76.7 & 51.6 & 15.1  & 66.7 &
           17.5  & 45.4 & 71.8 & 82.4 & 32.6 & 42.9 & 71.9 & 53.3 & 60.9  & 65.2 & 53.8  \\
Kantorov \etal\cite{kantorov2016}$^{*}$
         & 83.3  & 68.6 & 54.7 & 23.4 & 18.3 & 73.6 & 74.1 & 54.1 & 8.6   & 65.1 &
      47.1  & 59.5 & 67.0 & 83.5 & 35.3 & 39.9 & 67.0 & 49.7 & 63.5  & 65.2 & 55.1 \\
Diba \etal\cite{diba2016weakly}$^{*}$
         & 83.9  & 72.8 & 64.5 & 44.1 & 40.1 & 65.7 & 82.5 & 58.9 & 33.7  & 72.5 &
           25.6  & 53.7 & 67.4 & 77.4 & 26.8 & 49.1 & 68.1 & 27.9 & 64.5  & 55.7 & 56.7 \\
Zhu~\etal\cite{zhu2017soft}$^{*}$
         & 85.3  & 64.2 & 67.0 & 42.0 & 16.4 & 71.0 & 64.7 & 88.7 & 20.7 & 63.8 & 58.0 &
           84.1  & 84.7 & 80.0 & 60.0 & 29.4 & 56.3 & 68.1 & 77.4 & 30.5 & 60.6 \\
Dong~\etal\cite{dong2017dual}$^{*}$
		 & 85.3  & 71.9 & 66.8 & 27.0 & 26.5 & 81.2 & 78.5 & 36.1 & 17.2 & 80.6 &
           61.8  & 76.1 & 86.3 & 83.6 & 22.2 & 43.6 & 74.8 & 60.6 & 67.6 & 70.5 & 60.9 \\
Tang~\etal\cite{tang2017multiple}$^{*}$
         & 85.8  & 82.7 & 62.8 & 45.2 & 43.5 & 84.8 & 87.0 & 46.8 & 15.7 & 82.2 &
           51.0  & 45.6 & 83.7 & 91.2 & 22.2 & 59.7 & 75.3 & 65.1 & 76.8 & 78.1 & 64.3 \\
Teh~\etal\cite{tehattention}$^{*}$
         & 84.0  & 64.6 & 70.0 &62.4 & 25.8 & 80.6 & 73.9 & 71.5 & 35.7 & 81.6 &
           46.5  & 71.2 & 79.1 & 78.8 & 56.7 & 34.3 & 69.8 & 56.7 & 77.0 & 72.7 & 64.6 \\\hline
SPL+Fast R-CNN
         & 63.3  & 72.3 & 49.6 & 43.8 & 42.4 & 54.4 & 78.7 & 58.1 & 35.4 & 72.8 &
           43.0  & 63.1 & 78.1 & 82.3 & 59.1 & 37.8 & 68.8 & 56.6 & 64.5 & 51.7 & 58.8$\pm$0.7 \\
SPL+R-FCN
         & 39.2  & 54.8 & 59.0 & 38.6 & 34.5 & 53.7 & 73.7 & 62.2 & 36.2 & 73.6 &
           8.0   & 61.8 & 75.1 & 78.9 & 57.1 & 22.1 & 75.5 & 45.5 & 67.9 & 47.4 & 53.2$\pm$1.2 \\
SPL+Ensemble
         & 54.6  & 65.0 & 71.2 & 50.8 & 52.1 & 62.4 & 81.9 & 67.7 & 41.4 & 74.5 &
           21.0  & 69.6 & 78.4 & 86.5 & 66.5 & 46.1 & 76.0 & 57.6 & 74.7 & 56.3 & 62.7$\pm$0.9 \\
\textbf{MSPLD}
         & 66.0  & 71.2 & 67.9 & 49.7 & 52.9 & 68.8 & 82.6 & 76.6 & 42.5 & 81.6 &
           24.0  & 75.5 & 78.4 & 89.0 & 62.0 & 33.1 & 79.2 & 58.5 & 78.9 & 71.1 & \BF{65.5$\pm$0.3} \\
\hline
\end{tabular}

\label{table:corloc_voc2007train}
\end{table*}

{\bf Discussion of model convergence.}
Algorithm~\ref{alg:more_spld} adopts the AOS to solve MSPLD.
It alternatively updates the parameters of the object detectors and the parameters of the regularization terms.
When updating the parameters for the regularization terms, we can achieve the optimal solution via Eq.~\eqref{eq:update_v}.
When updating the parameters for the object detectors (CNN models), the model should converge to a local minimum by loss back propagation.
This alternative updating procedure converges when all the unlabeled samples have been traversed and when the objective function in Eq.~\eqref{eq:more_spld} cannot be further minimized.
Therefore, the algorithm will finally converge.

{\bf Model complexity.}
Suppose the time complexity of training a detector is $O(Flops)$, where Flops represents the floating-point operations of the network forward procedure. The overall time complexity of MSPLD then relies on the number of iterations in the alternative optimization strategy and the number of detectors. Based on Algorithm~\ref{alg:more_spld}, the time complexity of MSPLD is $O(iter_{max}\times m\times Flops)$, where the $m$ is the number of detectors and $iter_{max}$ is the maximum iteration number.
On PASCAL VOC'07, MSPLD can converge in no more than six iterations, and the standard setting of MSPLD may take about 50 hours using one GTX 1080 Ti GPU on PASCAL VOC. To learn new concept, we need to change the structure of the last classification layer and bounding box regression layer of the detectors. Therefore, we need to re-train the model based on the new data.

\section{Experimental Evaluation}

In this section, we compare MSPLD with some baselines on several large object detection benchmark datasets at first.
Secondly, we analysis the effect of different aspects of MSPLD to demonstrate the performance contribution of each composition in MSPLD.
Thirdly, we show the impact of supervision level in our algorithm by using different annotation information.
Lastly, with the visualized error analysis, we show how to further improve MSPLD in the future.

\subsection{Datasets}
We evaluate our method on PASCAL VOC 2007\cite{everingham2010the}, PASCAL VOC 2012~\cite{everingham2015the}, MS COCO 2014~\cite{lin2014microsoft}, and ILSVRC 2013 detection dataset~\cite{russakovsky2015imagenet}.
These four datasets are the most widely used benchmarks in the object detection task.
PASCAL VOC 2007 contains 10022 images annotated with bounding boxes for 20 object categories.
It is officially split into 2501 training, 2510 validation, and 5011 testing images.
PASCAL VOC 2012 is similar to PASCAL VOC 2007, but contains more images: 5717 training, 5823 validation images and 10991 testing images.
MS COCO 2014 contains 80k images for training and 40k images for validation, which are categorized into 80 classes.
ILSVRC 2013 is a large dataset with 200 categories for the detection task, which contains more than 400k images.
The standard training, validation and test splits for training and evaluation are used for these three datasets.

\begin{table}[!t]
\centering
\setlength{\tabcolsep}{2.5pt}
\caption{
Performance comparison on PASCAL VOC 2007 of different proposal generation methods.
}
\begin{supertabular}{|c|c|c|c|}                                    \hline
       & Selective Search & EdgeBox & Selective Search + EdgeBox   \\\hline
mAP    & 41.7             & 39.5    & \BF{41.9}                    \\\hline
CorLoc & 65.5             & 65.2    & \BF{65.6}                    \\
\hline
\end{supertabular}
\label{table:voc_proposal}
\end{table}

\subsection{Implementation Details}

\textbf{The details of detection models.}
We build R-FCN and Fast R-CNN on various base models as different detection models.
Three base models are tested in our experiments
\footnote{We suggest the following two-fold standards to select models in our method. First, each selected single model should exhibit possibly good performance in object detection. Second, the selected models should be possibly different from each other in aspects such as model structure and training strategy. In this manner, these models will be largely complementary to each other to guide a good performance of the final performance.}, i.e., GoogleNet~\cite{szegedy2015going}, VGG~\cite{simonyan2014very}, and ResNet~\cite{he2015deep}.
These models are pre-trained on ILSVRC 2012~\cite{krizhevsky2012imagenet}.
A boosting method, i.e., online hard example mining (OHEM)~\cite{shrivastava2016training}, is also tested in our experiments to study the complementarity between different models.
Region proposals are extracted by SS~\cite{uijlings2013selective} using the fast version or EB~\cite{zitnick2014edge},
following the standard practice used in \cite{Bilen16,dai2016r,girshick2015fast}.
We extract about 2000 proposals using SS and EB, respectively.
Proposals are extracted by SS in most experiments by default.
When we use both SS and EE (denoted SS+EE) to extract proposals, the total generated proposals are about 4000 for each image.
We use ImageNet pre-trained models to make a fair comparison with other algorithms~\cite{wang2014weakly,zhangbridging2016,tehattention,diba2016weakly}, because they also utilize ImageNet pre-trained models to provide a good initialization.

\textbf{Hyper-parameters.}
We do not tune the parameter $\gamma^{j_1,j_2}$ and always set it to $0.2/(m-1)$ in all our experiments for simplicity.
In our experiments, for the $c$-th class, $\lambda_{c}$ is decided by the number of selected images $|V_{c}|$, where $V_{c}$ is the $c$-th column of $V$.
In fact, according to Eq.~\eqref{eq:update_v}, a specific $\lambda_{c}$ corresponds to a specific $V_{c}$.
Moreover, given a specific $V_{c}$, there is one $\lambda_{c}$ that can correspond to such $V_{c}$. Therefore, we can use $|V_{c}|$ instead of $\lambda_{c}$ to compute $V_{c}$ in Eq.~\eqref{eq:update_v}.
In the implementation process of our experiments, supposing the number of selected images for the $c$-th class is $R_{k}=|V_{c}|$ at the $k$-th iteration, then this number will increase to $\frac{R_{k}(k+1)}{k}$ at the ($k$+1)-th iteration.
At the first iteration, $R_{k}$ is the initial number of labeled images for each class.
During basic detector training, we set the total training epochs to nine.
We empirically use the learning rate 0.001 for the first eight epochs and reduce it to 0.0001 for the last epoch.
In addition, the momentum and weight decay are set to 0.9 and 0.0005, respectively.
The first two convolution layers of each network are fixed, following~\cite{dai2016r,girshick2015fast}.
We randomly flip the image for data augmentation in the training phase.

\begin{table}[!t]
\centering
\caption{
Performance comparison on the PASCAL VOC 2012, MS COCO 2014, and ILSVRC 2013 datasets.
On PASCAL VOC 2012, mAP is evaluated on the test set and CorLoc is evaluated on the trainval set.
On ILSVRC 2013, we show the detection performance on the validation set.
On MS COCO 2014, we use the location prediction mAP for evaluation, following the same setting in~\cite{oquab2015object}.
}
\subfloat[PASCAL VOC 2012]{
\label{table:pascal_voc2012}
\begin{tabular}{|c|c|c|} \hline
  Methods                                  &      mAP       &    CorLoc     \\\hline
Li~\etal\cite{li2016weakly}                &      29.1      &      -        \\
Kantorov~\etal\cite{kantorov2016}          &      35.3      &     54.8      \\
Diba~\etal\cite{diba2016weakly}            &    \BF{37.9}   &      -        \\  \hline
\textbf{MSPLD}                             &      35.4      &   \BF{64.6}   \\
\hline
\end{tabular}
}
\hspace{0.6cm}
\subfloat[ILSVRC 2013]{
\label{table:imagenet}
\begin{tabular}{|c|c|} \hline
Methods                                         &       mAP        \\\hline
Wang~\etal\cite{wang2014weakly}                 &       6.0        \\
Felzenszwalb~\etal\cite{felzenszwalb2010object} &       8.8        \\
Li~\etal\cite{li2016weakly}                     &       10.8       \\
Diba~\etal\cite{diba2016weakly}                 &     \BF{16.3}    \\  \hline
\textbf{MSPLD}                                  &       13.9       \\
\hline
\end{tabular}
}
\subfloat[MS COCO 2014]{\label{table:coco}
\begin{tabular}{|c|c|} \hline
Methods                                         &       mAP     \\\hline
Oquab~\etal\cite{oquab2015object}               &      41.2     \\
Sun~\etal\cite{sun2016pronet}                   &      43.5     \\
Bency~\etal\cite{bency2016weakly}               &      47.9     \\
Zhu~\etal\cite{zhu2017soft}                     &      55.3     \\\hline
\textbf{MSPLD}                                  &      \BF{56.6}     \\
\hline
\end{tabular}
}
\label{table:large_scale}
\end{table}

\textbf{Evaluation metrics.}
Average precision (AP) is used on the testing data to evaluate detection accuracy;
correct localization (CorLoc)~\cite{deselaers2012weakly} is calculated for the training data to evaluate localization accuracy;
the location prediction mAP is calculated for the validation data to evaluate location prediction accuracy, following~\cite{oquab2015object}.
We use an intersection-over-union (IoU) ratio of 50\% for CorLoc and leverage the official evaluation code provided by~\cite{everingham2010the} to calculate AP.

\textbf{Initially labeled images.}
For each class, we randomly label $k$ images, which contain the box for this class.
We use $k=3$ if not specified, which results in a total of 60 initial annotated images. All the object bounding boxes in these 60 images are annotated, so in effect there are an average of 4.2 images per class, since some images have multiple classes.

\subsection{Comparison with State-of-the-art Algorithms}

We compare MSPLD with recent state-of-the-art WSOD algorithms
\cite{Bilen16,li2016weakly,kantorov2016,tehattention,wang2014weakly,diba2016weakly,zhangbridging2016}.
Fair comparisons are claimed because many of these methods use multiple models as well.
Bilen \etal \cite{Bilen16} use ensembles to improve performance.
Li~\etal~\cite{li2016weakly} use multiple steps.
They first train a classification model and apply a MIL model to mine the confident objects,
and then fine-tune a detection model to detect the objects.
Diba \etal \cite{diba2016weakly} cascade three networks: a location network, a segmentation network and a MIL network, and apply multi-scale data argumentation.
`SPL+Ensemble' in Table 2/3 represents the late fusion of multiple models.
This method simply averages the confidence scores and the refined bounding boxes (Eq.~\eqref{eq:cnn_out}), then follows the standard NMS and thresholding procedures.
In our comparison, we present the best results from their articles.
To evaluate the sensitivity of our method \emph{w.r.t} different initialization, we use random seeds to generate different initial fully annotated images.
For each experiment, we repeat four times, and mean performance and the standard deviation are reported.
Even if we only use few strong annotations for each class, our fused detection model can reduce the sensitivity to the initial annotated images.

\begin{table}[!t]
\centering
\setlength{\tabcolsep}{8pt}
\caption{
The performance of each detector employed in MSPLD.
``MV' indicates the use of multi-modal learning.
``w/o MV'' indicates we use the traditional self-paced method without multi-modal learning.
}
\begin{supertabular}{|l|c|c|c|} \hline
      Models                            &  Eval.  &  MV       & w/o MV  \\\hline
\multirow{2}{*}{Fast R-CNN (VGG16)}     &  mAP    &  36.0     & 33.7     \\
                                        &  CorLoc &  60.9     & 58.8     \\\hline
\multirow{2}{*}{R-FCN (Res50$^{ohem}$)} &  mAP    &  37.4     & 29.9     \\
                                        &  CorLoc &  62.7     & 53.2     \\\hline
\multirow{2}{*}{R-FCN (Res101)}         &  mAP    &  38.3     & 31.4     \\
                                        &  CorLoc &  62.0     & 54.1     \\
\hline
\end{supertabular}
\label{table:co_train}
\end{table}

\textbf{Comparisons \emph{w.r.t.} AP.}
Table~\ref{table:map_voc2007test} summarizes the AP on the PASCAL VOC 2007 test set.
The competing methods usually use full image-level labels.
In contrast, we use the same set of images but with much fewer annotations: totally 60 annotated images and the others are free-labeled.
Although the annotated images account for less than 1\% of the total number of training images, MSPLD achieves 41.7\% mAP, a competitive performance compared to state-of-the-art WSOD algorithms.
Our results achieve the best performance on some specific classes, e.g., the AP of person, bottle and cat exceeds the second best by  16\%, 10\%, and 12\%, respectively.
We view \cite{zhangbridging2016} as a comparable baseline to our method, which leverages the same base model VGG16 as our ``SPL+Fast R-CNN'' baseline.
In comparison, our baseline method, \emph{SPL+Fast R-CNN}, uses fewer annotations, but outperforms \cite{zhangbridging2016} by 2.4\% and 10.3\% in mAP and CorLoc, respectively.
The \emph{SPL+Fast R-CNN} model is superior to \emph{SPL+R-FCN}, because Fast R-CNN may pay more attention to the pseudo boxes selection and thus benefits more from the SPL strategy.
However, the two different architectures complement each other well, demonstrated by the improved performance of the \emph{SPL+Ensemble}.
Further, the proposed MSPLD is superior to the multi-model ensemble.
From Table~\ref{table:map_voc2007test} and Table~\ref{table:corloc_voc2007train}, MSPLD outperforms the model ensemble method by about 4\% in mAP and 3\% in CorLoc.
This observation further validates the effectiveness of our multi-model learning strategy.

\begin{table}[!t]
\setlength{\tabcolsep}{1.9pt}
\centering
\caption{
Ablation studies.
``\#Models'' represents the number detection models used.
``R-'' indicates the R-FCN detector, and ``F-'' indicates the Faster RCNN detector.
``R50'', ``VGG16'', ``Gog'', and ``R101'' indicate the base models, ResNet-50, VGG-16, GoogleNet-v1, and ResNet-101, respectively.
``ohem'' indicates whether the OHEM module is embedded.
``no prior'' represents that the filtration strategy is not used.
``no SPL'' means that we directly train the model with all the data after filtration, rather than using SPL.
}
\begin{tabular}{|c|l|c|c|}
\hline
\#Models              & Detection Model                            & mAP  & CorLoc  \\\hline
\multirow{6}{*}{1}    & R-R50 no prior                             & 28.6 & 50.1    \\
                      & R-R50 no SPL                               & 27.2 & 44.7    \\
                      & R-R50                                      & 28.9 & 50.6    \\
                      & R-R50$^{ohem}$                             & 29.9 & 53.2    \\
                      & R-Gog$^{ohem}$                             & 24.9 & 50.6    \\
                      & F-VGG16 no prior                           & 32.8 & 60.1    \\
                      & F-VGG16                                    & 33.7 & 60.9    \\\hline
\multirow{3}{*}{2}    & R-R50$^{ohem}$ + F-VGG16                   & 38.3 & 63.4    \\
                      & R-R50$^{ohem}$ + R-Gog$^{ohem}$            & 32.1 & 57.3    \\
                      & R-Gog$^{ohem}$ + F-VGG16                   & 35.8 & 61.6    \\\hline
\multirow{5}{*}{3}    & R-R50$^{ohem}$ + F-VGG16 + R-Gog$^{ohem}$  & 38.5 & 62.8    \\
                      & R-R50$^{ohem}$ + F-VGG16 + R-R101          & \BF{41.7} & \BF{65.5}    \\
                      & R-R50$^{ohem}$ + F-VGG16 + R-R101$^{ohem}$ & 38.9 & 63.4    \\
                      & R-R50$^{ohem}$ + F-VGG16$^{ohem}$ + R-R101 & 37.5       & 61.4    \\
                      & R-R50$^{ohem}$ + F-VGG16$^{ohem}$ + R-R101$^{ohem}$ & 37.1  & 61.1\\
\hline
\end{tabular}
\label{table:ablation}
\end{table}

\textbf{Comparisons \emph{w.r.t.} CorLoc.}
Table~\ref{table:corloc_voc2007train} shows the correct localization on the PASCAL VOC 2007 trainval set.
MSPLD achieves an average CorLoc 65.5\%, which sets a new state-of-the-art.
Note that \cite{tehattention} has a similar CorLoc to our MSPLD, but we obtain a much higher mAP than~\cite{tehattention} (41.7\% vs. 34.5\%).
From Table~\ref{table:map_voc2007test} and Table~\ref{table:corloc_voc2007train},
it can be seen that our method does not have large performance deviations under different initializations of fully annotated images.
Moreover, it can be observed from the tables that when using multiple models, the performance of our method is less sensitive to different initializations than that of the baseline single model.
\DXYB{
In Table~\ref{table:map_voc2007test} and Table~\ref{table:corloc_voc2007train}, we note that recent works~\cite{tang2017multiple,GeYY18} report very competitive accuracy with ours.
Their methods work under the traditional weakly-supervised setting, while our method is implemented under semi-supervised learning setting with only very few examples provided..
Specifically, Tang~\etal\cite{tang2017multiple} and Ge~\etal\cite{GeYY18} achieve higher mAP than MSPLD, and MSPLD achieves higher CorLoc than~\cite{tang2017multiple}.
Two reasons may contribute to their higher mAP.
First, they use superior architectures to generate region proposals rather than the selective search method in our work.
Second, they employ multi-scale training strategies, but we use a single-scale training strategy.
The advantage of our work over~\cite{tang2017multiple,GeYY18} is that we are able to make better use multiple models to improve the performance of a single model.
}

{\bf Results on large-scale datasets.}
Table~\ref{table:pascal_voc2012} presents the mAP and CorLoc of MSPLD on PASCAL VOC 2012, which also achieves the competitive performance compared with others.
We also compared our algorithm on ILSVRC 13 only with \cite{wang2014weakly,felzenszwalb2010object,li2016weakly,diba2016weakly}, since no other weakly supervised or few shot algorithms have been tried on this dataset.
Results on Table~\ref{table:imagenet} are similar to the previous one, we achieves the competitive performance with fewer annotation informations on ILSVRC 2013 validation set.
Following~\cite{oquab2015object}, Table~\ref{table:coco} uses the location prediction~\cite{oquab2015object} mean average precision to compare our results with others on MS COCO 2014.
As shown in Table~\ref{table:large_scale}, our algorithm achieves competitive or superior results on the large-scale detection datasets.

\textbf{Comparison of different variants.}
We compare the impact of different proposal generations methods.
SS, EB and their combination are tested.
The results are presented in Table \ref{table:voc_proposal}.
We find that EB is inferior to SS due to its poorer initialization in the first iteration.
Combining both of the two region proposals, we obtain a slight performance improvement.

\textbf{The effect of multi-modal learning.}
Furthermore, we demonstrate the performance of the individual detection models with and without multi-modal learning in Table~\ref{table:co_train}.
The displayed models are used with \emph{MSPLD} shown in Table~\ref{table:map_voc2007test}.
We observe that the performance of individual detection models is much higher when using multi-modal learning, which proves the effectiveness of our method in enhancing each model.

\subsection{Ablation Studies}
We examine the contribution of different components of MSPLD on PASCAL VOC 2007 and MS COCO 2014.

\textbf{The impact of different models and the curriculum regime $\Psi$.}
From Table~\ref{table:ablation}, several conclusions can be made.
(1) Since \emph{R-R50} outperforms \emph{R-R50 no SPL} and \emph{R-R50 no prior}, we prove that the data selection strategy and prior knowledge are necessary.
(2) Fast R-CNN with VGG16 achieves the best single model performance.
(3) We observe that \emph{R-R50} and \emph{F-VGG16} are complementary and benefit from the multi-modal learning.
The reason may be that R-FCN has the position-sensitive layer for box refinement, while Fast R-CNN with VGG-16 focuses more on the proposals' classification.
(4) The use of \emph{ohem} sightly improves mAP for \emph{R-R50}, but harms the performance of \emph{F-VGG16} and \emph{R-R101}.
(5) When adding \emph{ohem} to \emph{R-R101} or to \emph{F-VGG16}, we observe inferior results.
The probable reason for this observation is that VGG16 and ResNet-101 are larger than ResNet-50 and that the training set is relatively small (in our few-example setting). Therefore, the influence of \emph{ohem} on VGG-16 and ResNet-101 is limited or even negative.

\begin{table}[t]
\centering
\caption{
Performance comparison of MSPLD on PASCAL VOC 2007 using different numbers of noisy images for the MSPLD model with $k=3$ for initialization.
}
\begin{supertabular}{|c|c|c|c|c|c|} \hline
   \#noisy images  &  0     &  1000  &  2000  & 5000 & 10000   \\\hline
   noise scale     &  0\%   &  20\%  &   40\% & 100\%&  200\%  \\\hline
    mAP            &  41.7  &  39.9  &  39.8  & 39.8 &  39.3   \\
    CorLoc         &  65.5  &  64.3  &  64.0  & 63.9 &  63.5   \\
\hline
\end{supertabular}
\label{table:noise}
\end{table}

\begin{figure}[t]
\centering\includegraphics[width=\linewidth]{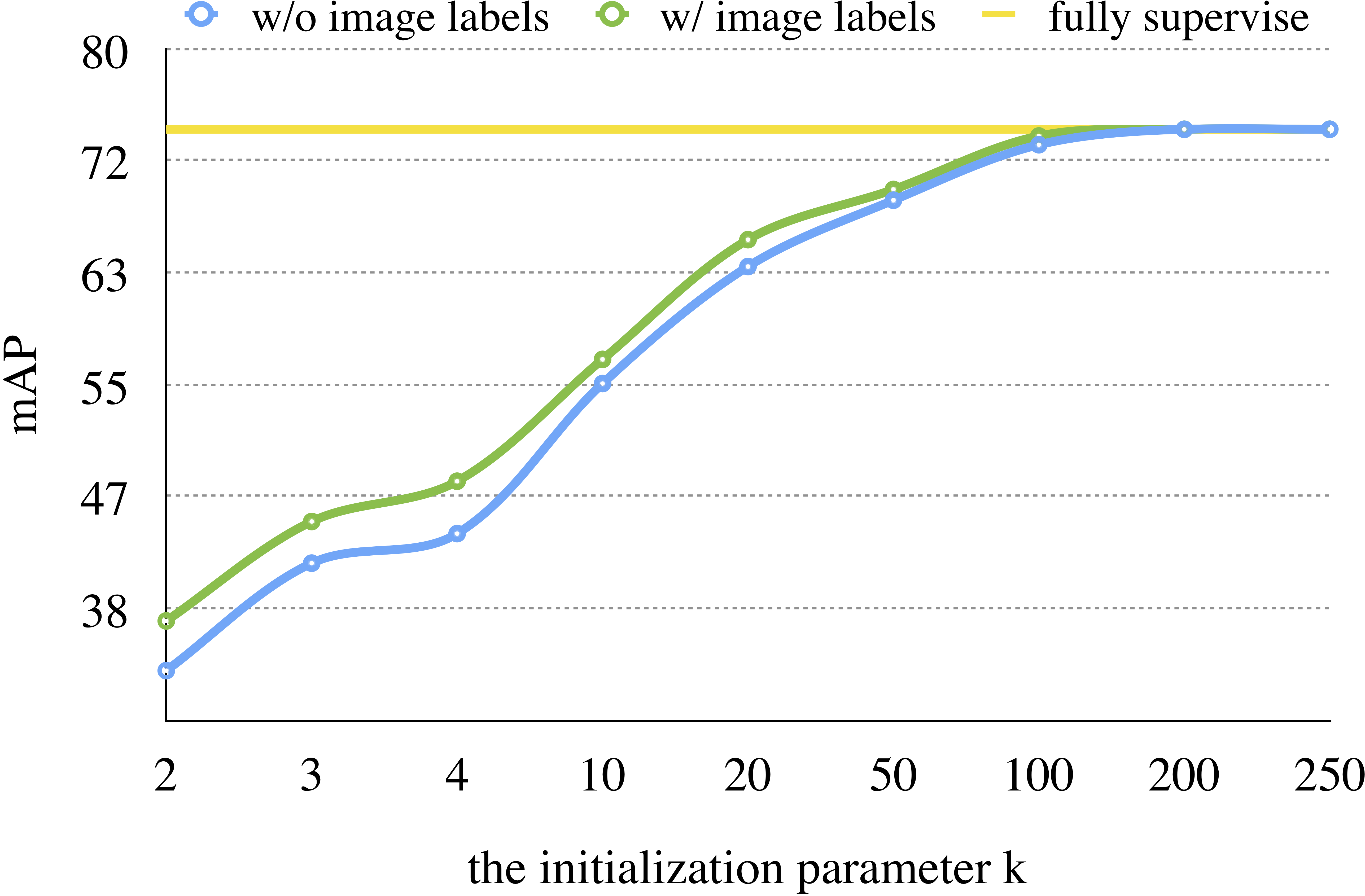}
\caption{
Performance comparison of MSPLD on PASCAL VOC 2007 using different selection numbers for the initial labeled images.
In ``w/ image label'', we simply leverage the image label to filter the undesired pseudo boxes.
}
\label{fig:initialization}
\end{figure}

\textbf{The impact of the number of initial labels.}
Using $k=2$ (totally 40 images in PASCAL VOC 2007) for initialization is not stable for training, and can result into severely reduced accuracy.
We can observe that even one additional example per class could significantly improve the performance of our MSPLD.
In Figure~\ref{fig:initialization}, each category has a maximum of 250 images on average, which can reproduce a fully supervised object detector~\cite{girshick2015fast,dai2016r}.
In our method, when 100 images are randomly selected during initialization, we can obtain very close accuracy to the fully-supervised method.
In this paper, we choose to use only 3-4 images which will suffice to ensure a decent accuracy at little manual cost.

\textbf{The impact of image-level labels.}
Image-level supervision can be easily incorporated into our framework.
We use the simplest approach to embed this supervision, i.e., only using the image label to filter out incorrect pseudo boxes.
The results are shown in Figure~\ref{fig:initialization}.
The simplest method for appending image-level labels can greatly boost our framework.

\textbf{The robustness regarding the noisy images.}
All previous experiments are based on well-annotated datasets.
For example, we know that the images in PASCAL VOC 2007 contain at least one object of the 20 classes.
Therefore, we have added images from YFCC100M~\cite{thomee2016yfcc100m} as noisy images to the PASCAL VOC 2007 dataset.
This experiment can make our algorithm completely unsupervised and demonstrate its robustness against outliers.
Specifically, we first randomly sampled 10,000 images from YFCC100M and used various numbers of images from these 10,000 images as noisy images.
We then employed this augmented dataset for detector learning.
Results are shown in Table 8.
It can be observed that our approach still yields a competitive detection accuracy when more than half of the augmented dataset are noisy images. These results demonstrate the robustness of our method against outliers.

\begin{figure}[t]
\centering\includegraphics[width=\linewidth]{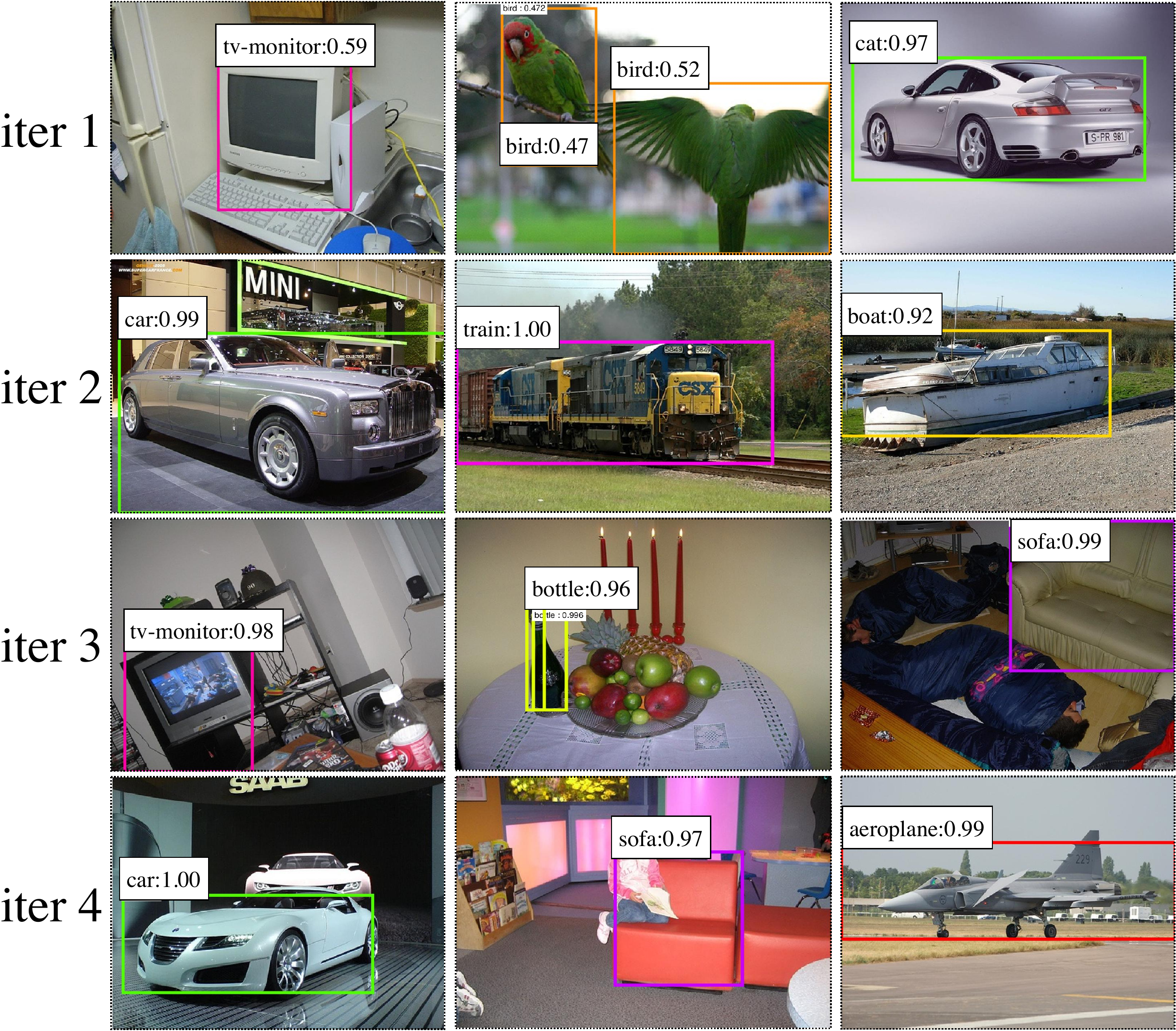}
\caption{
Qualitative results of MSPLD over the training iterations.
The boxes with different colors indicate the generated pseudo boxes by our method for different classes.
}
\label{fig:sample_over_training}
\end{figure}

\textbf{Analysis of the generalization ability.}
Since all the classes of the detection datasets are contained in the 1000 classes of the ImageNet dataset, the pre-trained models use some pre-knowledge of their detection classes.
Such knowledge may benefit the quality of the detectors obtained by MSPLD.
To demonstrate the generalization ability of MSPLD, we use pre-trained models that are not trained on the detection classes.
To this end, we construct Non-overlapping ImageNet-VOC/COCO sets for pre-training.
For PASCAL VOC 2007, we manually select 746 ImageNet classes, which do not overlap with the 20 detection classes of PASCAL VOC. Images from these selected 746 classes compose of the None-overlap ImageNet-VOC subset.
For MS COCO 2014, we manually select 706 ImageNet classes, which do not overlap with the 80 detection classes of MS COCO. Image samples from the selected 706 classes form the None-overlap ImageNet-COCO subset.
We use such constituted Non-overlapping ImageNet-VOC and Non-overlapping ImageNet-COCO sets to pre-train the VGG16, ResNet-50, and ResNet-101 models for experiments on PASCAL VOC and MS COCO 2014, respectively.
We observe that mAP on PASCAL VOC 2007 drops from 41.7\% to 38.2\%; the localization prediction mAP on MS COCO 2014 drops from 56.6\% to 53.3\%.
There might be two reasons that cause such performance drop.
The first should be the lack of detection classes during pre-training, while another important reason should be the less number of pre-trained data.
To evaluate which one causes the performance drop, we have randomly sampled 74.6\% training images from ImageNet to form the Overlapping ImageNet-VOC set, which contains the same number of training data with the Non-overlapping ImageNet-VOC set, but is not enforced not to contain PASCAL VOC classes
We then use Overlapping ImageNet-VOC to pre-train the VGG16, ResNet-50, and ResNet-101 models for experiments on PASCAL VOC. We observe that mAP on PASCAL VOC 2007 drops from 41.7\% to 38.9\%. The performance of Overlapping ImageNet-VOC pre-training is almost similar to the performance with Non-Overlapping ImageNet-VOC.
This verifies that pre-training without the detection classes does not substantially affect the performance of MSPLD.

\subsection{Qualitative Analysis}

\textbf{Qualitative results over the training iterations.}
We show pseudo-labeled images by MSPLD over the training iterations in Figure~\ref{fig:sample_over_training}.
Briefly, in the first iteration, the detector tends to choose images with relatively high classification confidence aggregated over the bounding boxes.
After the detector is updated, it can gradually label objects in more complicated situation, e.g., the rotated TV monitor and several small bottles in Figure~\ref{fig:sample_over_training}.

\begin{figure}[t]
\centering\includegraphics[width=\linewidth]{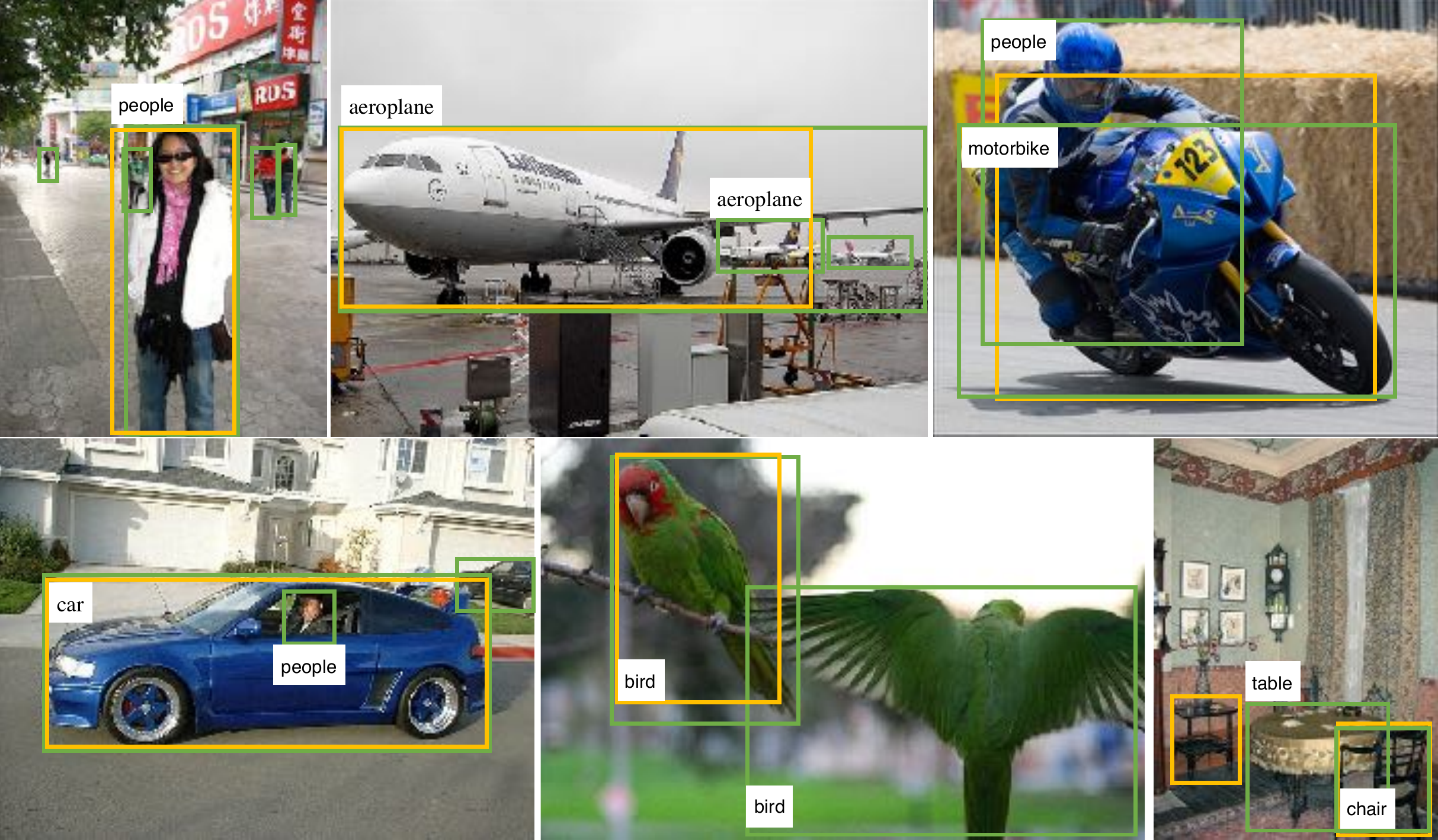}
\caption{
{\bf Qualitative results} of the {\bf inaccurate} pseudo instance-level labels generated by MSPLD during the training procedure.
The green boxes indicate the ground-truth object annotation.
The yellow boxes indicate the generated pseudo boxes by MSPLD.
The white blocks show the object classes.
}
\label{fig:wrong_sample}
\end{figure}

\textbf{Error analysis.}
Some of the images that are newly generated by our method are shown in Figure~\ref{fig:wrong_sample}.
We observe that the generated pseudo boxes have the good localization accuracy, but cannot detect every object in complex images.
For example, the pseudo boxes correctly localize the true objects in the first five images.
However, all these images contain multiple objects, and have occlusions, or overlaps between the objects.
The generated boxes do not cover all objects well, which will compromise the performance of the final detectors.
Prior knowledge could filter out some of the complex images, but this problem remains to be solved.
We will focus on generating robust pseudo boxes for complex images in the future.

\section{Conclusion and Future Work}
In this paper, we propose an object detection framework (MSPLD) that uses only a few bounding box labels per category by consistently implementing iterations between detector amelioration and reliable sample selection.
To enhance its detector learning capability with the scarcity of annotation, MSPLD embeds multiple detection models in its learning scheme.
It can fully use the discriminative knowledge for different detection models, and possibly complement them to ameliorate the detector training quality.
Under such extremely limited supervision information, MSPLD can achieve competitive performance compared to state-of-the-art WSOD approaches, which use more supervised knowledge of samples than our method.

MSPLD still requires about 1\% of the images in the entire dataset to be annotated.
In future, we will focus on further reducing the annotation information, i.e., only using one image per class, to obtain the similar performance.
Except for the improvement of the base CNN feature and the object detector, the challenges are how to initialize the detector from limited annotation and, design a robust learning scheme to ameliorate the detector stably.
Besides, we will investigate to improve our method into accommodating novel classes while simultaneously not destroy the accuracy of the training models on the previously trained ones.

\bibliographystyle{IEEEtran}
\bibliography{IEEEabrv,egbib}

\begin{IEEEbiography}[{\includegraphics[width=1in,height=1.25in,clip,keepaspectratio]{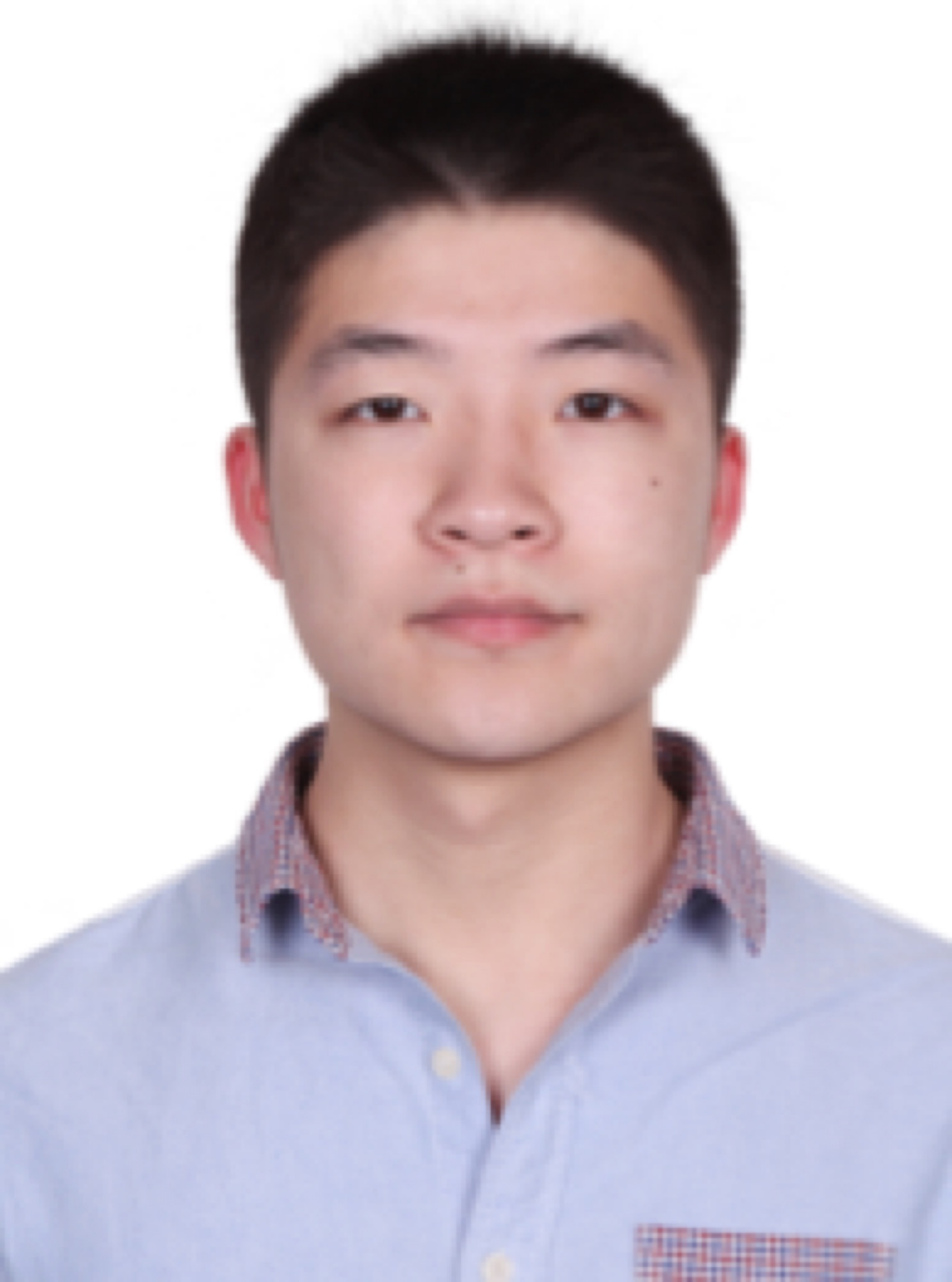}}]{Xuanyi Dong} received the B.E. degree in Computer Science and Technology from Beihang University, Beijing, China, in 2016.
He is currently a Ph.D. student in the Center of Artificial Intelligence, University of Technology Sydney, Australia, under the supervision of Prof. Yi Yang.
\end{IEEEbiography}

\begin{IEEEbiography}[{\includegraphics[width=1in,height=1.25in,clip,keepaspectratio]{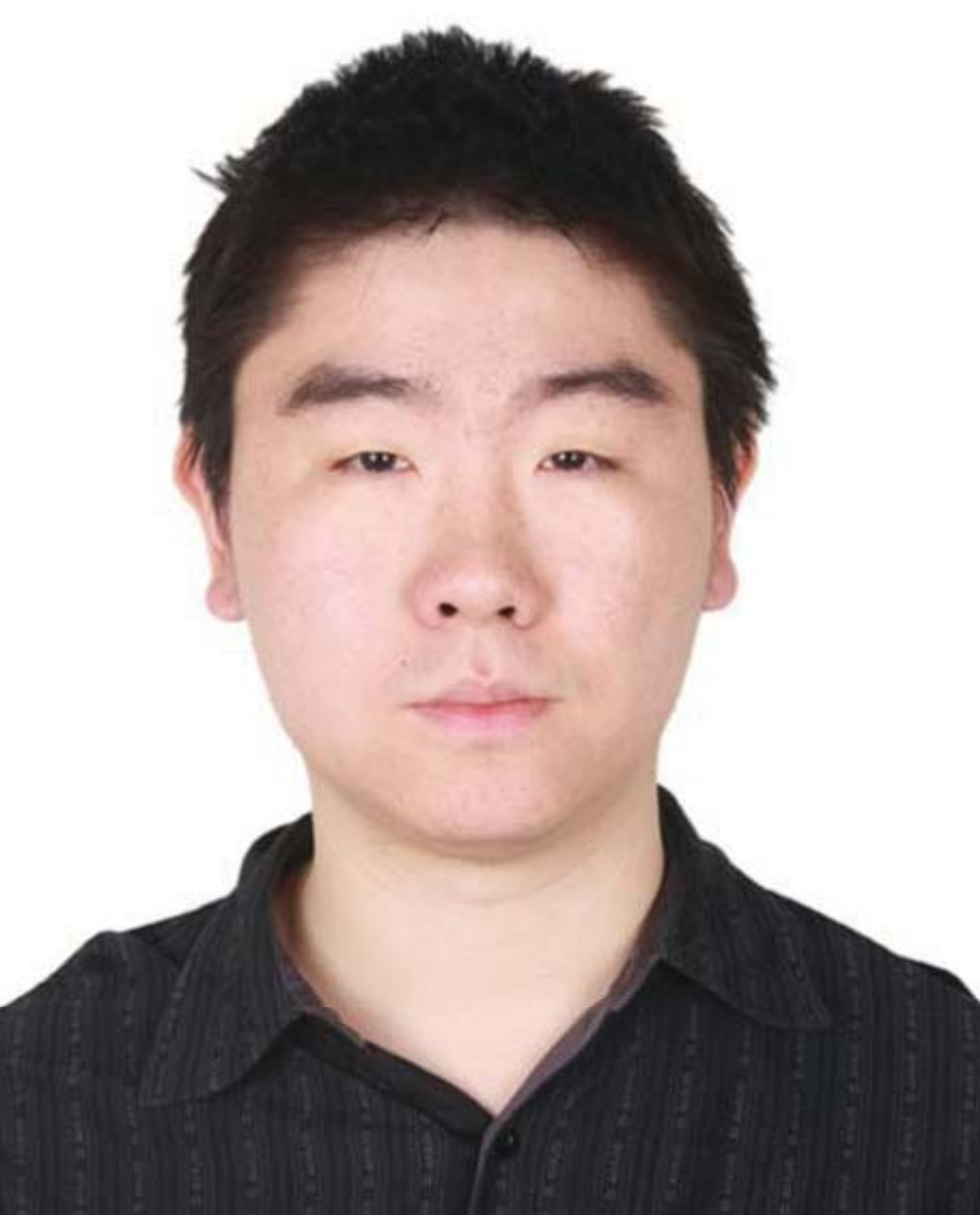}}]{Liang Zheng}
received the Ph.D degree in Electronic Engineering from Tsinghua University, China, in 2015, and the  B.E. degree in Life Science from Tsinghua University, China, in 2010. He was a postdoc researcher in University of Texas at San Antonio, USA. He is now a postdoc researcher in the Center of Artificial Intelligence,  University of Technology Sydney, Australia. His research interests are image retrieval, person re-identification and deep learning.
\end{IEEEbiography}

\begin{IEEEbiography}[{\includegraphics[width=1in,height=1.25in,clip,keepaspectratio]{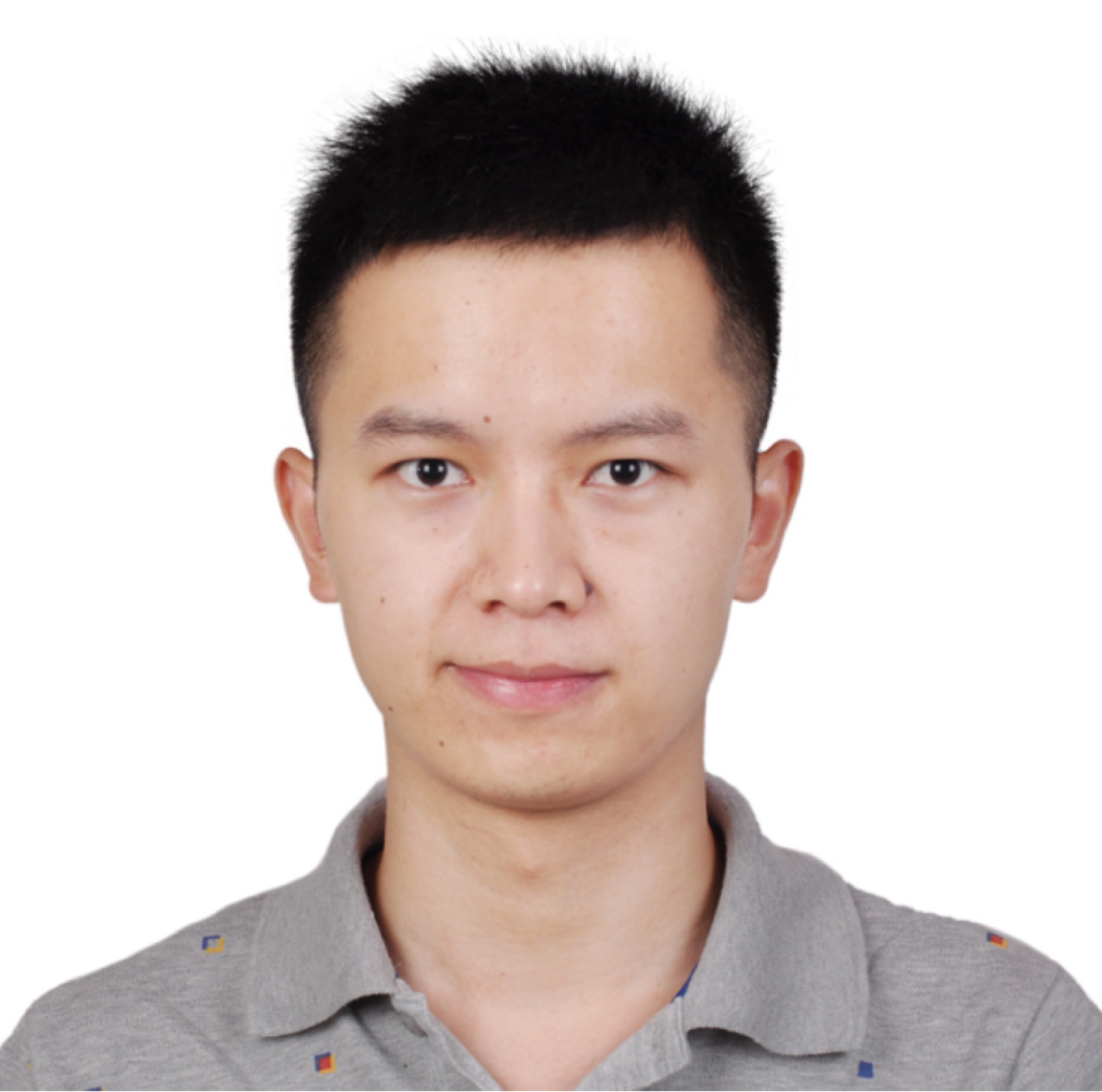}}]{Fan Ma} received the B.Sc. and M.Sc degree from Xi'an Jiaotong University, Xi'an, China, in 2014 and 2017, respectively, under the supervisoin of Prof. Deyu Meng. He is currently a Ph.D. student in the Center of Artificial Intelligence, University of Technology Sydney, Australia, under the supervision of Prof. Yi Yang. His research interests include machine learning and computer vision, especially on semi-supervised learning, self-paced learning and person re-identification.
\end{IEEEbiography}

\begin{IEEEbiography}[{\includegraphics[width=1in,height=1.25in,clip,keepaspectratio]{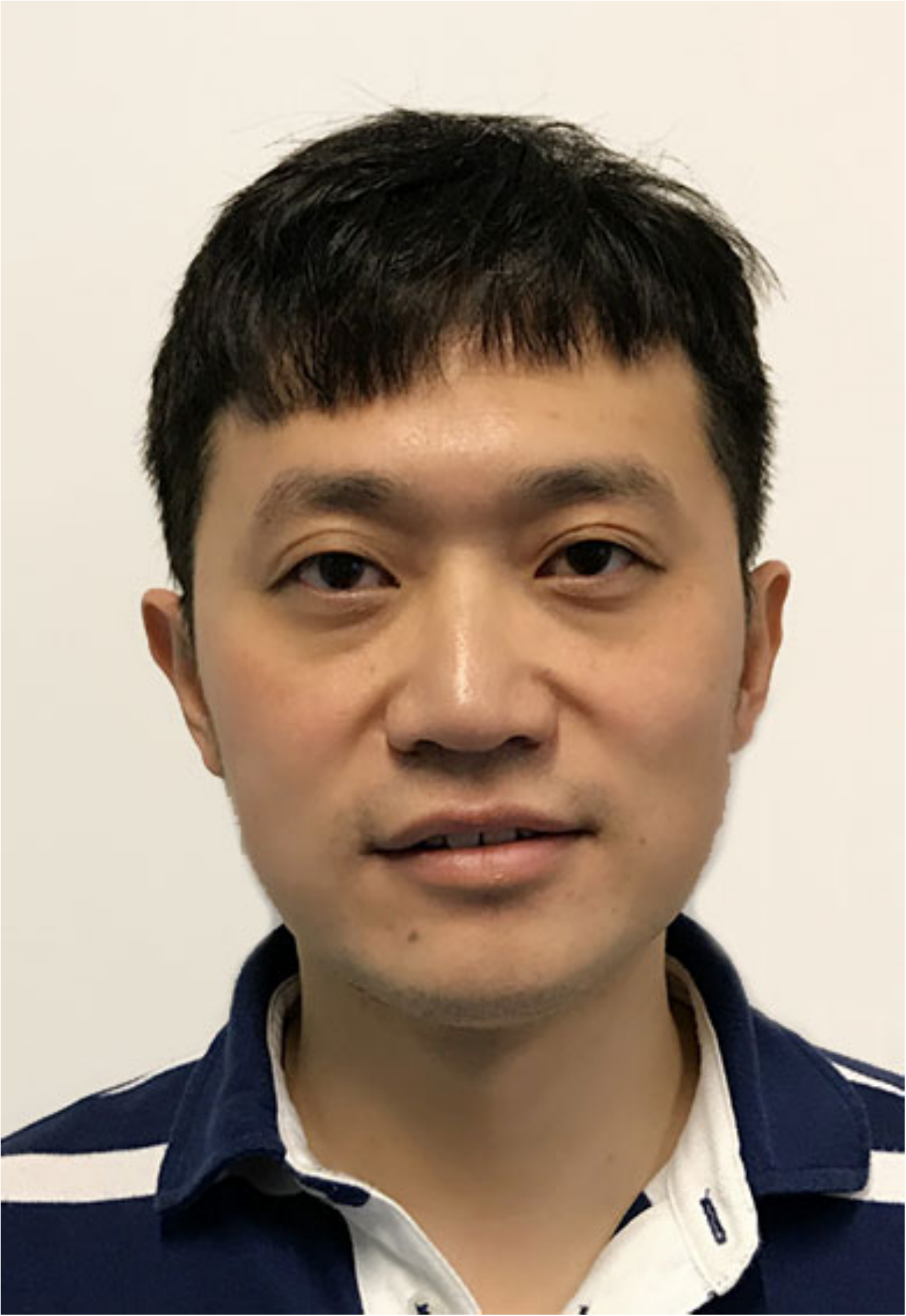}}]{Yi Yang} received the Ph.D. degree in computer
science from Zhejiang University, Hangzhou, China, in 2010.
He was a Post-Doctoral Research with the School of Computer Science, Carnegie Mellon University, Pittsburgh, PA, USA.
He is currently a Professor with University of Technology Sydney, Australia.
His current research interest includes machine learning and its applications to multimedia content analysis and computer vision.
\end{IEEEbiography}

\begin{IEEEbiography}[{\includegraphics[width=1in,height=1.25in,clip,keepaspectratio]{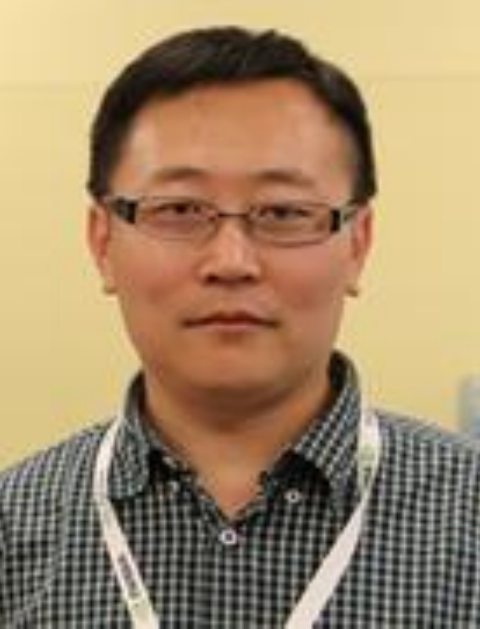}}]{Deyu Meng} received the B.Sc., M.Sc., and Ph.D.
degrees from Xi'an Jiaotong University, Xi'an,
China, in 2001, 2004, and 2008, respectively.
He is currently a Professor with
the Institute for Information and System Sciences,
School of Mathematics and Statistics, Xi'an Jiaotong
University. From 2012 to 2014, he took his two-
year sabbatical leave in Carnegie Mellon University.
His current research interests include self-paced
learning, noise modeling, and tensor sparsity.
\end{IEEEbiography}

\end{document}